\documentclass[letterpaper, 10 pt, conference]{ieeeconf}  

\usepackage{amsmath,amsfonts}
\usepackage{algorithmic}
\usepackage{array}
\usepackage[caption=false,font=normalsize,labelfont=sf,textfont=sf]{subfig}
\usepackage{textcomp}
\usepackage{stfloats}
\usepackage{pifont}
\usepackage{cite}
\makeatletter
\let\NAT@parse\undefined
\makeatother
\usepackage{lineno,hyperref}
\usepackage{booktabs} 
\hypersetup{
	colorlinks=true, 
	linkcolor=blue,  
	citecolor=blue, 
	urlcolor=red     
}
\usepackage{url}
\usepackage{verbatim}
\usepackage{graphicx}
\usepackage{xcolor}
\usepackage{amsmath,booktabs,makecell}
\def\BibTeX{{\rm B\kern-.05em{\sc i\kern-.025em b}\kern-.08em
		T\kern-.1667em\lower.7ex\hbox{E}\kern-.125emX}}
\usepackage{balance}

\title{\LARGE \bf
Reduced Cartesian Kinetostatics for Tendon-Driven Continuum Robots: Residual-Stabilized Full-Shape Propagation}

\author{Ke Wu$^{1}$, Fangju Yang$^{1,2}$, Xiaohui Zhang$^{3}$, Zhengqiang Zhang$^{2}$, Jingang Yi$^{4}$, Jian S. Dai$^{5}$
}

\begin{document}

\maketitle
\thispagestyle{empty}
\pagestyle{empty}

\begin{abstract}
Many planning and control tasks for tendon-driven continuum robots (TDCRs) require the complete Cartesian backbone geometry. We present a reduced Cartesian framework for planar, axially compressible TDCRs that propagates equilibrium configurations along prescribed tendon-force and tendon-displacement trajectories. The backbone is represented by two global position fields. Following exact variation, a Taylor--Galerkin reduction condenses prescribed spatial properties and distributed loads into offline moment vectors, yielding analytic reduced residuals and Jacobians without online spatial quadrature or numerical differentiation. Analytical differentiation and residual correction yield first-order rate systems requiring one fixed-dimensional linear solve per rate evaluation after initial equilibrium alignment on a regular branch. Across four simulated cases covering variable tendon routing, nonuniform geometry, axial compression, and their combined effects, the propagated Cartesian shapes and distributed strains closely match pointwise geometric variable-strain (GVS) equilibrium solutions. Residual correction suppresses propagation drift across the tested step sizes while adding only about $0.98\,\%$ to the mean update time of uncorrected Euler. The proposed method requires $0.508\,\mathrm{ms}$ per update on average, approximately $11$ times faster than pointwise GVS solves. Displacement-driven experiments yield a maximum normalized mean backbone position error of $1.02\,\%$ and a maximum end-effector position error of $0.28\,\%$. These results support efficient and accurate Cartesian full-shape prediction along prescribed actuation paths.

\end{abstract}



\section{Introduction}

Forward kinetostatics is fundamental to the design
\cite{rao2021model,cao2017workspace}, planning \cite{wang2024quasistatic}, and
model-based control \cite{della2023model} of tendon-driven continuum
robots (TDCRs). It characterizes how tendon actuation and external loading
determine the equilibrium configuration and, consequently, the robot's
workspace and task performance
\cite{camarillo2008mechanics,rucker2011statics,rao2021model}.
Unlike rigid-link mechanisms, continuum robots deform continuously
along their backbones, making the choice of configuration variables
central to the formulation of their equilibrium mechanics
\cite{armanini2023soft,liu2003approach}.
Existing kinetostatic models therefore differ substantially in how they
represent the backbone configuration. 

At the kinematic level, piecewise-constant-curvature (PCC) models
represent each section by a few arc variables and recover the
backbone pose geometrically
\cite{webster2010design,jones2006kinematics}. Mechanics-based models
instead take rotation, strain, or Cartesian position fields as their primary unknowns. Wu et al.~\cite{wu2026lightweight} parameterize the cross-sectional rotation field and derive closed-form kinetostatic solutions. Cosserat rod models parameterize distributed strain fields \cite{rucker2011statics,rodriguez2007origami}, while geometric variable-strain
models approximate them in finite bases whose coefficients serve as generalized coordinates \cite{renda2020geometric}. These fields naturally express rod constitutive laws and elastic energy but require spatial integration to recover the Cartesian backbone. Cartesian formulations instead choose global position or displacement fields as unknowns and derive strains from their spatial derivatives. Spline,
finite-element, and absolute-state models respectively use Cartesian NURBS control points \cite{luo2020spline}, nodal displacements \cite{bieze2018finite}, and Cartesian nodal positions augmented by orientation variables \cite{sadati2019reduced}. Thus, strain coordinates align with rod mechanics, whereas Cartesian position coordinates place task-level geometry directly in the equilibrium solution.

The chosen representation changes the form of the equations, but
geometric nonlinearity generally leaves a nonlinear equilibrium problem
\cite{gilbert2021mathematical}. Cosserat formulations solve nonlinear
spatial equations subject to boundary conditions
\cite{rucker2011statics}; geometric variable-strain formulations reduce
the distributed mechanics to nonlinear equilibrium equations in strain
coordinates \cite{renda2020geometric}; and finite-element formulations
solve projected nodal systems through constrained numerical
optimization \cite{bieze2018finite}. Computational benchmarks show that
solution time and convergence depend on the model resolution and
initial guess; during deployment or trajectory tracking, the preceding
shape can therefore be reused to initialize the next equilibrium solve
\cite{rao2021model}. Quasistatic Jacobian methods further differentiate
implicit equilibrium conditions to obtain local
actuation-to-configuration sensitivities for continuum-robot motion
planning \cite{greigarn2019jacobian}. Continuation methods have also
been used to trace equilibrium branches of magnetically actuated
continuum robots \cite{peyron2018magnetic}. Recent quasi-static planning
characterizes stable continuum-robot configurations as a smooth
implicit manifold and constructs local tangent charts, while projecting
each sampled configuration onto equilibrium through a direct-shooting
boundary-value solve \cite{wang2024quasistatic}. Together, these studies expose the differential structure of
continuum-robot equilibrium sets through local sensitivity analysis,
branch continuation, and tangent-chart construction.


Motivated by these observations, this paper recasts TDCR forward
kinetostatics from successive nonlinear equilibrium solves into
residual-stabilized tracking of the complete Cartesian backbone along
actuation-parameterized equilibrium branches. The main contributions
are summarized as follows:

\begin{enumerate}

\item \textbf{Task-aligned Cartesian kinetostatics.}
We formulate the planar, axially compressible kinetostatic problem in the two global Cartesian components of the backbone position field, making the complete equilibrium shape a direct Cartesian output of the kinetostatic model.

\item \textbf{Offline-precomputable Galerkin reduction.}
We develop a post-variational Taylor--Galerkin reduction that projects the continuous spatial equilibrium problem onto a compact polynomial coefficient space and condenses prescribed spatial properties and load
profiles into offline moment vectors, retaining their distributed effects while improving computational efficiency.

\item \textbf{Explicit residual-stabilized kinetostatic propagation.}
We analytically differentiate the reduced equilibrium equations and
incorporate residual correction to obtain explicit first-order rate
systems for Cartesian full-shape propagation along regular equilibrium
branches, requiring one fixed-dimensional linear solve per rate
evaluation. Predictions closely match pointwise geometric
variable-strain (GVS) solutions, while experiments yield a maximum
normalized mean backbone position error of $1.02\%$ and a maximum
normalized end-effector position error of $0.28\%$. The method requires
$0.508~\mathrm{ms}$ per update on average, approximately $11\times$
faster than pointwise GVS solves, while suppressing the residual
drift observed with uncorrected Euler propagation.

\end{enumerate}
\section{Problem Statement}

\subsection{The studied Manipulator}

\begin{figure}[!t]
\centering
\includegraphics[width=0.92\columnwidth]{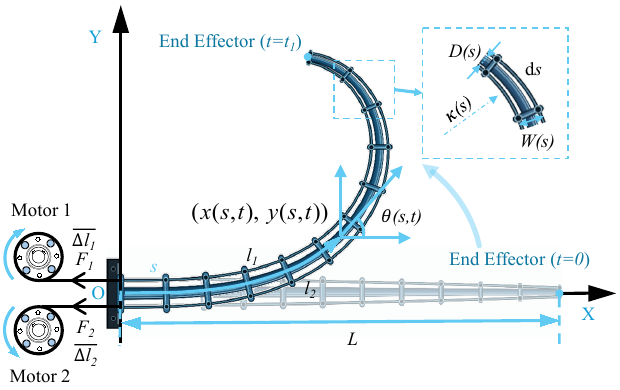}
\caption{The studied tendon-driven continuum robot.}
\label{fig:studied-manipulator}
\end{figure}

As shown in Fig.~\ref{fig:studied-manipulator}, we consider
a planar, axially compressible tendon-driven continuum robot with a
straight and stress-free reference configuration. Let \(s\in[0,L]\)
denote the reference arc-length coordinate of the backbone. Two tendons
are routed longitudinally on opposite sides of the backbone. With
tendon friction neglected, the prescribed actuation inputs are either
the spatially uniform tendon tensions \(F_1(t),F_2(t)\geq0\) or the tendon
displacements
\(\overline{\Delta l}_1(t),\overline{\Delta l}_2(t)\).  The prescribed axial and bending stiffnesses \(EA(s)\) and \(EI(s)\),
together with the tendon-routing diameter \(W(s)\), may vary along the
backbone. The robot may also be subjected to prescribed signed distributed loads \(q_x(s)\) and \(q_y(s)\), defined per unit reference length and acting in the fixed global Cartesian directions.

\subsection{Objective}

The objective is to establish a computationally efficient Cartesian
kinetostatic model that continuously propagates the complete
equilibrium configuration along prescribed tendon-force or
tendon-displacement trajectories while accounting for axial compression,
bending, external loading, and spatially varying mechanical and
geometric properties.

\section{Methodology}
\subsection{Cartesian Variational Formulation}
\label{sec:cartesian-variational-formulation}

Let
\(\mathcal F=\{O;\hat{\mathbf x},\hat{\mathbf y}\}\)
denote a fixed global Cartesian frame. The origin \(O\) is placed at
the robot base, \(\hat{\mathbf x}\) is aligned with the stress-free
backbone, and \(\hat{\mathbf y}\) defines the transverse direction.
The backbone centerline expressed in \(\mathcal F\) is
\([x(s),y(s)]^{\mathbf T}\). The clamped-base conditions are
\begin{equation}
\begin{aligned}
x(0)&=0,
&
y(0)&=0,
&
y_s(0)&=0.
\end{aligned}
\label{eq:base-conditions}
\end{equation}
A subscript \(s\) denotes differentiation with respect to \(s\).
Admissible configurations satisfy
\(x_s^2+y_s^2>0\) over \(s\in[0,L]\), with \(x_s(0)>0\).
Define the geometric scalar
\(\chi=x_s y_{ss}-y_s x_{ss}\).
The axial stretch ratio and bending strain are
\begin{equation}
\lambda=\sqrt{x_s^2+y_s^2},
\ \ 
\kappa=\frac{\chi}{\lambda^2}.
\label{eq:cartesian-strains}
\end{equation}
In particular, \(\kappa\) is the rotation gradient per unit reference length. Following the approximate tendon-displacement formulation in \cite{yang2026lightweightC}, the two tendon displacements are
\begin{equation}
\begin{aligned}
\Delta l_1
&=
-\int_0^L
\left(
\lambda-1-\frac{W(s)}{2}\kappa
\right)\mathrm ds,
\\
\Delta l_2
&=
-\int_0^L
\left(
\lambda-1+\frac{W(s)}{2}\kappa
\right)\mathrm ds.
\end{aligned}
\label{eq:tendon-displacements}
\end{equation}
For prescribed tendon forces, combining the tendon-induced potential
work with the axial and bending energies and the distributed-load
potentials gives the total potential energy
\begin{equation}
\begin{aligned}
\Pi[x,y]
=&
\int_0^L\Bigg[
\frac{EA(s)}{2}(\lambda-1)^2
+\frac{EI(s)}{2}\kappa^2
\\
&+(F_1+F_2)(\lambda-1)
-\frac{W(s)}{2}(F_1-F_2)\kappa
\\
&-q_x(s)(x-s)-q_y(s)y
\Bigg]\mathrm ds .
\end{aligned}
\label{eq:total-potential}
\end{equation}
The equilibrium configuration follows from the stationarity condition
\(\delta\Pi=0\). The variations of the two strain measures required to
evaluate this condition are
\begin{equation}
\begin{aligned}
\delta\lambda
&=
\frac{x_s\delta x_s+y_s\delta y_s}{\lambda},
\\
\delta\kappa
&=
\frac{
y_{ss}\delta x_s-x_{ss}\delta y_s
-y_s\delta x_{ss}+x_s\delta y_{ss}}
{\lambda^2}
\\
&\quad
-\frac{2\kappa}{\lambda^2}
\left(
x_s\delta x_s+y_s\delta y_s
\right).
\end{aligned}
\label{eq:strain-variations}
\end{equation}
Substituting these expressions into the variation of
\eqref{eq:total-potential} yields
\begin{equation}
\begin{aligned}
\delta\Pi
=
&\int_0^L\Bigg\{
\left[
EA(s)(\lambda-1)+F_1+F_2
\right]\delta\lambda
\\
&+
\left[
EI(s)\kappa
-\frac{W(s)}{2}(F_1-F_2)
\right]\delta\kappa
\\
&-q_x(s)\delta x-q_y(s)\delta y
\Bigg\}\mathrm ds .
\end{aligned}
\label{eq:first-variation}
\end{equation}
For compactness, let \(\psi\) denote the integrand of
\eqref{eq:total-potential}. Collecting the coefficients of
\(\delta x_s\), \(\delta y_s\), \(\delta x_{ss}\), and
\(\delta y_{ss}\) gives
\begin{equation}
\begin{aligned}
\frac{\partial\psi}{\partial x_s}
={}&
\left[
EA(s)(\lambda-1)+F_1+F_2
\right]\frac{x_s}{\lambda}
\\
&+
\left[
EI(s)\kappa
-\frac{W(s)}{2}(F_1-F_2)
\right]
\frac{y_{ss}-2\kappa x_s}{\lambda^2},
\\[1mm]
\frac{\partial\psi}{\partial y_s}
={}&
\left[
EA(s)(\lambda-1)+F_1+F_2
\right]\frac{y_s}{\lambda}
\\
&-
\left[
EI(s)\kappa
-\frac{W(s)}{2}(F_1-F_2)
\right]
\frac{x_{ss}+2\kappa y_s}{\lambda^2},
\\[1mm]
\frac{\partial\psi}{\partial x_{ss}}
={}&
-
\left[
EI(s)\kappa
-\frac{W(s)}{2}(F_1-F_2)
\right]
\frac{y_s}{\lambda^2},
\\[1mm]
\frac{\partial\psi}{\partial y_{ss}}
={}&
\left[
EI(s)\kappa
-\frac{W(s)}{2}(F_1-F_2)
\right]
\frac{x_s}{\lambda^2}.
\end{aligned}
\label{eq:potential-derivatives}
\end{equation}
Integrating the first-derivative terms once and the second-derivative
terms twice transfers all spatial derivatives from the admissible
variations. Since the interior variations \(\delta x\) and
\(\delta y\) are arbitrary, \(\delta\Pi=0\) gives
\begin{equation}
\begin{aligned}
\frac{\mathrm d}{\mathrm ds}
\left(
\frac{\partial\psi}{\partial x_s}
\right)
-
\frac{\mathrm d^2}{\mathrm ds^2}
\left(
\frac{\partial\psi}{\partial x_{ss}}
\right)
+q_x(s)
&=0,
\\
\frac{\mathrm d}{\mathrm ds}
\left(
\frac{\partial\psi}{\partial y_s}
\right)
-
\frac{\mathrm d^2}{\mathrm ds^2}
\left(
\frac{\partial\psi}{\partial y_{ss}}
\right)
+q_y(s)
&=0.
\end{aligned}
\label{eq:cartesian-equilibrium}
\end{equation}
The boundary terms generated by the same integration-by-parts
operation provide the natural conditions. Then, the free-tip force boundary conditions are
\begin{equation}
\small
\begin{aligned}
\left[
\frac{\partial\psi}{\partial x_s}
-\frac{\mathrm d}{\mathrm ds}
\left(
\frac{\partial\psi}{\partial x_{ss}}
\right)
\right]_{s=L}=0,\ 
\left[
\frac{\partial\psi}{\partial y_s}
-\frac{\mathrm d}{\mathrm ds}
\left(
\frac{\partial\psi}{\partial y_{ss}}
\right)
\right]_{s=L}
=0.
\end{aligned}
\label{eq:tip-force-conditions}
\end{equation}
The two boundary terms associated with the tip slope variations are
linearly dependent and reduce to the single moment condition
\begin{equation}
\begin{aligned}
\left[
EI(s)\kappa
-\frac{W(s)}{2}(F_1-F_2)
\right]_{s=L}
&=0.
\end{aligned}
\label{eq:tip-moment-condition}
\end{equation}
At the base, the only unconstrained slope variation is
\(\delta x_s(0)\). Its boundary term vanishes identically because
\(y_s(0)=0\) implies
\(\left.\partial\psi/\partial x_{ss}\right|_{s=0}=0\).

\subsection{Post-Variational Galerkin Reduction}
\label{sec:post-variational-galerkin}
Substitution of \eqref{eq:cartesian-strains} into
\eqref{eq:potential-derivatives} introduces the non-polynomial metric
factors $\lambda^{-2m}$. These
factors retain the unknown coefficients inside reciprocal powers,
preventing their separation from the Galerkin integrals. Therefore,
after obtaining the exact field equations and boundary conditions, we
approximate
\begin{equation}
\begin{aligned}
\lambda^{-2m}
\approx
\mathcal T_m^{(P)}
=
\sum_{p=0}^{P}
\binom{-m}{p}
\left(\lambda^2-1\right)^p,\ m\in\left\{\frac{1}{2},1,2,3\right\}.
\end{aligned}
\label{eq:taylor-kernel}
\end{equation}
The expansion is centered at the stress-free metric. Since only
nondegenerate axial compression is considered,
\begin{equation}
0<\lambda\leq1
\quad\Longrightarrow\quad
\left|\lambda^2-1\right|
=
1-\lambda^2
<1.
\label{eq:taylor-compression-domain}
\end{equation}
The continuation is restricted to configurations satisfying
\(0<\lambda(s)\leq1\) over \(s\in[0,L]\).
Hence, the underlying binomial series converges throughout the
considered compressive regime, while the finite-order approximation
becomes less accurate as the compression increases. The same order $P$ is used in the field equations and natural boundary
conditions. For polynomial order \(n\), the centerline is approximated by
\begin{equation}
\begin{aligned}
x(s)
\approx
s+\sum_{i=1}^{n}a_i s^i,\  y(s)
\approx
\sum_{i=2}^{n}b_i s^i,
\\
\mathbf c
=
\begin{bmatrix}
a_1,\ldots,a_n,b_2,\ldots,b_n
\end{bmatrix}^{\mathbf T}
\in\mathbb R^{2n-1}.
\end{aligned}
\label{eq:cartesian-polynomial}
\end{equation}
Together, the finite Taylor truncation and the Cartesian polynomial
ansatz render the variational terms polynomial in $s$, enabling
coefficient separation and spatial precomputation. The geometric quantity \(\lambda^2-1\) has the polynomial expansions
\begin{equation}
\begin{aligned}
\lambda^2-1
\approx {}&
2\sum_{i=1}^{n}i a_i s^{i-1}
+\sum_{i=1}^{n}\sum_{j=1}^{n}
ij\,a_i a_j s^{i+j-2}
\\
&+\sum_{i=2}^{n}\sum_{j=2}^{n}
ij\,b_i b_j s^{i+j-2},
\end{aligned}
\label{eq:geometric-polynomials}
\end{equation}
and similarly \(\chi\) has the following
\begin{equation}
\begin{aligned}
\chi
\approx {}
\sum_{j=2}^{n}j(j-1)b_j s^{j-2}+\sum_{i=1}^{n}\sum_{j=2}^{n}
ij(j-i)a_i b_j s^{i+j-3}.
\end{aligned}
\label{eq:chi-polynomial}
\end{equation}
To retain the origin of each physical contribution, collect the known
spatial functions as $\mathbf f(s)=
\begin{bmatrix}
EA(s) & EI(s) & 1 & W(s)
\end{bmatrix}^{\mathbf T}$. The four entries correspond, respectively, to axial elasticity,
bending elasticity, common tendon actuation, and differential tendon
actuation. The truncated variational derivatives can then be written
uniformly as\begin{equation}
\begin{aligned}
\left[
\frac{\partial\psi}{\partial\zeta}
\right]_P
&=
\mathbf f^{\mathbf T}(s)
\mathbf g_{\zeta}^{(P)}
\left(s;\mathbf c,F_1,F_2\right),\ 
\zeta
\in
\left\{
x_s,y_s,x_{ss},y_{ss}
\right\}.
\end{aligned}
\label{eq:general-variational-function}
\end{equation}
The complete polynomial functions associated with the \(x\) direction
are
\begin{equation}
\small
\begin{aligned}
\mathbf g_{x_s}^{(P)}
={}&
\begin{bmatrix}
x_s\left(1-\mathcal T_{1/2}^{(P)}\right)
\\
\chi y_{ss}\mathcal T_{2}^{(P)}
-2\chi^2x_s\mathcal T_{3}^{(P)}
\\
(F_1+F_2)x_s\mathcal T_{1/2}^{(P)}
\\
(F_1-F_2)
\left[
-\dfrac{1}{2}y_{ss}\mathcal T_{1}^{(P)}
+\chi x_s\mathcal T_{2}^{(P)}
\right]
\end{bmatrix},
\\
\mathbf g_{x_{ss}}^{(P)}
={}&
\begin{bmatrix}
0
\\
-\chi y_s\mathcal T_{2}^{(P)}
\\
0
\\
\dfrac{1}{2}(F_1-F_2)y_s\mathcal T_{1}^{(P)}
\end{bmatrix}.
\end{aligned}
\label{eq:x-variational-functions}
\end{equation}
Similarly, the \(y\)-direction functions are
\begin{equation}
\small
\begin{aligned}
\mathbf g_{y_s}^{(P)}
={}&
\begin{bmatrix}
y_s\left(1-\mathcal T_{1/2}^{(P)}\right)
\\
-\chi x_{ss}\mathcal T_{2}^{(P)}
-2\chi^2y_s\mathcal T_{3}^{(P)}
\\
(F_1+F_2)y_s\mathcal T_{1/2}^{(P)}
\\
(F_1-F_2)
\left[
\dfrac{1}{2}x_{ss}\mathcal T_{1}^{(P)}
+\chi y_s\mathcal T_{2}^{(P)}
\right]
\end{bmatrix},
\\
\mathbf g_{y_{ss}}^{(P)}
={}&
\begin{bmatrix}
0
\\
\chi x_s\mathcal T_{2}^{(P)}
\\
0
\\
-\dfrac{1}{2}(F_1-F_2)x_s\mathcal T_{1}^{(P)}
\end{bmatrix}.
\end{aligned}
\label{eq:y-variational-functions}
\end{equation}
Thus, \(\mathcal T_{1/2}^{(P)}\),
\(\mathcal T_{1}^{(P)}\),
\(\mathcal T_{2}^{(P)}\), and
\(\mathcal T_{3}^{(P)}\) are used explicitly in the axial, tendon, and
bending contributions. After substituting \eqref{eq:cartesian-polynomial}, each
\(\mathbf g_{\zeta}^{(P)}\) is a finite polynomial in \(s\):
\begin{equation}
\begin{aligned}
\mathbf g_{\zeta}^{(P)}
\left(s;\mathbf c,F_1,F_2\right)
&=
\sum_{\ell\geq0}
\mathbf g_{\zeta,\ell}^{(P)}
\left(\mathbf c,F_1,F_2\right)s^\ell ,
\end{aligned}
\label{eq:coefficient-functions}
\end{equation}
where only finitely many coefficient vectors
\(\mathbf g_{\zeta,\ell}^{(P)}\) are nonzero. These vectors are polynomial in \(\mathbf c\), affine in \(F_1\) and
\(F_2\), and independent of \(s\). All structural- and
actuation-related spatial integrations are collected into the
precomputed moment vectors
\begin{equation}
\small
\begin{aligned}
\mathbf C_k
&=
\int_0^L
\mathbf f(s)s^k\,\mathrm ds =
\begin{bmatrix}
\displaystyle\int_0^L EA(s)s^k\,\mathrm ds
\\[1mm]
\displaystyle\int_0^L EI(s)s^k\,\mathrm ds
\\[1mm]
\displaystyle\int_0^L s^k\,\mathrm ds
\\[1mm]
\displaystyle\int_0^L W(s)s^k\,\mathrm ds
\end{bmatrix}.
\end{aligned}
\label{eq:precomputed-spatial-vector}
\end{equation}
For prescribed robot properties, the required moment vectors
\(\mathbf C_k\) are computed once offline. For the reduction, we use the post-variationally truncated virtual-work
form corresponding to the exact equilibrium equations. Substituting the
admissible variations
$\delta x=s^r\delta a_r$ and
$\delta y=s^r\delta b_r$
yields the Bubnov--Galerkin residuals
\begin{equation}
\small
\begin{aligned}
R_{a_r}
={}&
r\sum_{\ell\geq0}
\left(
\mathbf g_{x_s,\ell}^{(P)}
\right)^{\mathbf T}
\mathbf C_{r+\ell-1}
\\
&+
r(r-1)\sum_{\ell\geq0}
\left(
\mathbf g_{x_{ss},\ell}^{(P)}
\right)^{\mathbf T}
\mathbf C_{r+\ell-2}
-
\int_0^L q_x(s)s^r\,\mathrm ds
=0,
\\
&\hspace{59mm}r=1,\ldots,n,
\\
R_{b_r}
={}&
r\sum_{\ell\geq0}
\left(
\mathbf g_{y_s,\ell}^{(P)}
\right)^{\mathbf T}
\mathbf C_{r+\ell-1}
\\
&+
r(r-1)\sum_{\ell\geq0}
\left(
\mathbf g_{y_{ss},\ell}^{(P)}
\right)^{\mathbf T}
\mathbf C_{r+\ell-2}
-
\int_0^L q_y(s)s^r\,\mathrm ds
=0,
\\
&\hspace{59mm}r=2,\ldots,n.
\end{aligned}
\label{eq:precomputed-galerkin-residuals}
\end{equation}
For \(r=1\), the second sum in \(R_{a_r}\) is omitted. Eq.
\eqref{eq:precomputed-galerkin-residuals} shows that the coefficient
functions are taken outside the spatial integrals. Since \(q_x(s)\) and
\(q_y(s)\) are prescribed, their Galerkin moments are likewise computed
once offline. Collecting the \(2n-1\) equations gives
\begin{equation}
\begin{aligned}
\mathbf R(\mathbf c,F_1,F_2)
&=
\begin{bmatrix}
R_{a_1},\ldots,R_{a_n},
R_{b_2},\ldots,R_{b_n}
\end{bmatrix}^{\mathbf T} 
\end{aligned}
\label{eq:reduced-equilibrium-model}
\end{equation}

\subsection{Explicit Residual-Stabilized Kinetostatic Propagation}
\label{sec:stabilized-kinetostatics}
Let \(t\) parameterize a differentiable prescribed actuation path,
let \(\mathbf c=\mathbf c(t)\), and let overdots denote differentiation
with respect to \(t\). Inertial and rate-dependent effects are excluded.

\subsubsection{Tendon-Force-Driven Kinetostatics}
Since the precomputed vectors \(\mathbf C_k\) are independent of
\(\mathbf c,F_1,F_2\), differentiation of
\eqref{eq:precomputed-galerkin-residuals} acts only on the coefficient
vectors. Hence,
\begin{equation*}
\begin{aligned}
\mathbf J
=
\frac{\partial\mathbf R}{\partial\mathbf c}\in
\mathbb R^{(2n-1)\times(2n-1)},\ 
\mathbf R_{F_i}
=
\frac{\partial\mathbf R}{\partial F_i}\in
\mathbb R^{2n-1},
\  i=1,2,
\end{aligned}
\end{equation*}
are evaluated analytically without online spatial quadrature or
numerical differentiation. Let \(F_1(t)\) and \(F_2(t)\) be
differentiable prescribed tendon-force trajectories. Since
\(\mathbf R=\mathbf R(\mathbf c,F_1,F_2)\),
\begin{equation*}
\begin{aligned}
\dot{\mathbf R}
=
\mathbf J\dot{\mathbf c}
+
\mathbf R_{F_1}\dot F_1
+
\mathbf R_{F_2}\dot F_2.
\end{aligned}
\end{equation*}
The algebraic condition $(\mathbf R=\mathbf0)$ defines the reduced equilibrium manifold. To correct residual drift during propagation, we
impose
$
\dot{\mathbf R}
+\gamma\mathbf R=
\mathbf0,
\ 
\gamma>0,
$
where \(\gamma\) specifies the residual-correction rate. On a regular equilibrium branch satisfying
\(\operatorname{rank}(\mathbf J)=2n-1\), the Jacobian is nonsingular.
Combining this total derivative with the above correction gives
\begin{equation}
\begin{aligned}
\dot{\mathbf c}
=
-\mathbf J^{-1}
\left(
\mathbf R_{F_1}\dot F_1
+
\mathbf R_{F_2}\dot F_2
+
\gamma\mathbf R
\right).
\end{aligned}
\label{eq:coefficient-rate-update}
\end{equation} The initial coefficients are aligned with the complete loading
condition by enforcing
\begin{equation}
\begin{aligned}
\mathbf R
\left(
\mathbf c(0),F_1(0),F_2(0)
\right)
&=
\mathbf0 .
\end{aligned}
\label{eq:initial-residual-alignment}
\end{equation}
With exact initial alignment, the continuous-time rate system preserves
\(\mathbf R=\mathbf0\). The correction term vanishes on the equilibrium
manifold and acts only when the propagated reduced state has a nonzero
residual.
For zero initial tendon forces and zero distributed loads, the
stress-free initialization is \(\mathbf c(0)=\mathbf0\). If tendon
pretension or distributed loads are initially present, \(\mathbf c(0)\)
is obtained from \eqref{eq:initial-residual-alignment} under the
corresponding initial loading condition. Eq. \eqref{eq:coefficient-rate-update} is integrated to update
\(\mathbf c\). The residual is then reevaluated from
\eqref{eq:reduced-equilibrium-model}, and the updated coefficients are
substituted directly into \eqref{eq:cartesian-polynomial} to recover the Cartesian equilibrium configuration.

\subsubsection{Tendon-Displacement-Driven Kinetostatics}
\label{sec:tendon-displacement-driven-kinetostatics}
The preceding formulation treats the tendon forces as prescribed
inputs. We now consider the alternative case in which
\(\overline{\Delta l}_1(t)\) and
\(\overline{\Delta l}_2(t)\) are differentiable prescribed
tendon-displacement trajectories. Since the actuation is
imposed kinematically, the tendon-force potential is removed, leaving
\begin{equation*}
\begin{aligned}
\Pi_0[x,y]
=
&\int_0^L
\Bigg[
\frac{EA(s)}{2}(\lambda-1)^2
+
\frac{EI(s)}{2}\kappa^2
\\
&-
q_x(s)(x-s)
-
q_y(s)y
\Bigg]\mathrm ds .
\end{aligned}
\end{equation*}
Using the tendon displacements in
\eqref{eq:tendon-displacements}, the prescribed inputs define the tendon-displacement constraints
\begin{equation}
\begin{aligned}
h_1[x,y;\overline{\Delta l}_1]
&=
\Delta l_1[x,y]
-
\overline{\Delta l}_1
=
0,
\\
h_2[x,y;\overline{\Delta l}_2]
&=
\Delta l_2[x,y]
-
\overline{\Delta l}_2
=
0.
\end{aligned}
\label{eq:prescribed-tendon-constraints}
\end{equation}
For compactness, the dependence of \(h_i\) and \(\mathbf h\) on the
prescribed tendon displacements is suppressed below. Introducing two auxiliary scalar multipliers
\(\eta_1\) and \(\eta_2\), the constrained functional is
\begin{equation}
\begin{aligned}
\widetilde{\Pi}
[x,y,\eta_1,\eta_2]
=
\Pi_0[x,y]
+
\eta_1h_1[x,y]
+
\eta_2h_2[x,y].
\end{aligned}
\label{eq:displacement-constrained-functional}
\end{equation}
The multipliers enforce the constrained stationarity and are treated
as auxiliary unknowns. The exact first variation is
\begin{equation}
\begin{aligned}
\delta\widetilde{\Pi}
=
\delta\Pi_0
+
\eta_1\delta h_1
+
\eta_2\delta h_2
+
h_1\delta\eta_1
+
h_2\delta\eta_2
=
0.
\end{aligned}
\label{eq:displacement-constrained-variation}
\end{equation}
Since the prescribed tendon displacements are fixed under variations
of \(x\) and \(y\), the corresponding constraint variations are
\begin{equation}
\small
\begin{aligned}
\delta h_1
&=
-\int_0^L
\left(
\delta\lambda
-
\frac{W(s)}{2}\delta\kappa
\right)\mathrm ds,
\\
\delta h_2
&=
-\int_0^L
\left(
\delta\lambda
+
\frac{W(s)}{2}\delta\kappa
\right)\mathrm ds.
\end{aligned}
\label{eq:exact-constraint-variations}
\end{equation}
Thus, variations with respect to \(x\) and \(y\) yield the exact
constrained mechanical virtual work, whereas the arbitrary variations
\(\delta\eta_1\) and \(\delta\eta_2\) recover the constraints
\(h_1=0\) and \(h_2=0\). Only after establishing this constrained variation do we apply
the post-variation Taylor truncation and Cartesian polynomial
approximation introduced in
Section~\ref{sec:post-variational-galerkin}. For the constraint
equations obtained from the multiplier variations, the required
truncated quantities are
\begin{equation}
\begin{aligned}
[\lambda-1]_P
=
\sum_{p=1}^{P}
\binom{1/2}{p}
\left(\lambda^2-1\right)^p,\ [\kappa]_P
=
\chi\mathcal T_1^{(P)}.
\end{aligned}
\label{eq:displacement-taylor-terms}
\end{equation}
Substituting the Cartesian polynomial approximation
\eqref{eq:cartesian-polynomial} gives the truncated displacement
constraints
\begin{equation}
\begin{aligned}
h_1^{(P)}(\mathbf c)
={}&
-\int_0^L
\left[
[\lambda-1]_P
-
\frac{W(s)}{2}[\kappa]_P
\right]\mathrm ds
-
\overline{\Delta l}_1,
\\
h_2^{(P)}(\mathbf c)
={}&
-\int_0^L
\left[
[\lambda-1]_P
+
\frac{W(s)}{2}[\kappa]_P
\right]\mathrm ds
-
\overline{\Delta l}_2.
\end{aligned}
\label{eq:truncated-tendon-constraints}
\end{equation}
These are global scalar constraints and therefore require no additional
Galerkin weighting. Collecting them gives
\begin{equation*}
\begin{aligned}
\mathbf h(\mathbf c)
=
\begin{bmatrix}
h_1^{(P)}(\mathbf c)
\\
h_2^{(P)}(\mathbf c)
\end{bmatrix}.
\end{aligned}
\end{equation*}
For the mechanical virtual work, the Taylor kernels are applied
directly to the exact expressions
\(\delta\Pi_0\), \(\delta h_1\), and \(\delta h_2\).
Substitution of the Cartesian polynomial and its admissible coefficient
variations then gives the Bubnov--Galerkin reductions
\begin{equation*}
\begin{aligned}
\left[\delta\Pi_0\right]_P
=
\delta\mathbf c^{\mathbf T}
\mathbf R_0(\mathbf c),
\ 
\mathbf R_0
\in
\mathbb R^{2n-1},
\end{aligned}
\end{equation*}
and
\begin{equation}
\begin{aligned}
\left[\delta h_1\right]_P
&=
\delta\mathbf c^{\mathbf T}
\mathbf B_1(\mathbf c),\ 
\left[\delta h_2\right]_P=
\delta\mathbf c^{\mathbf T}
\mathbf B_2(\mathbf c),
\\
\mathbf B(\mathbf c)
&=
\begin{bmatrix}
\mathbf B_1(\mathbf c)
&
\mathbf B_2(\mathbf c)
\end{bmatrix}
\in
\mathbb R^{(2n-1)\times2}.
\end{aligned}
\label{eq:displacement-multiplier-map}
\end{equation}
Here, \(\mathbf R_0\) contains the elastic and prescribed
distributed-load contributions, while the columns of \(\mathbf B\)
represent the coefficient-space variations of the two exact
tendon-displacement constraints. After the polynomial substitution, all integrands become finite
polynomials in \(s\). The displacement constraints require only the
moments
\(\int_0^L s^k\mathrm ds\) and
\(\int_0^L W(s)s^k\mathrm ds\), already contained in the third and
fourth entries of \(\mathbf C_k\) in
\eqref{eq:precomputed-spatial-vector}. Hence, the displacement-driven
formulation introduces neither online spatial quadrature nor an
additional precomputation structure. Let
\(
\boldsymbol{\eta}(t)
=
[\eta_1(t)\ \eta_2(t)]^{\mathbf T}
\).
Collecting the independently truncated mechanical and displacement
constraint equations gives
\begin{equation}
\begin{aligned}
\mathbf R_0(\mathbf c)
+
\mathbf B(\mathbf c)\boldsymbol{\eta}
&=
\mathbf0,\ \mathbf h(\mathbf c)=
\mathbf0.
\end{aligned}
\label{eq:reduced-displacement-system}
\end{equation}
For the subsequent continuation, define the Jacobian of the truncated
constraint values as
\begin{equation*}
\begin{aligned}
\mathbf H(\mathbf c)
=
\frac{\partial\mathbf h}{\partial\mathbf c}
\in
\mathbb R^{2\times(2n-1)}.
\end{aligned}
\end{equation*}
At finite Taylor order,
\(\mathbf B\) is intentionally kept distinct from
\(\mathbf H^{\mathbf T}\). The former follows from truncating the exact
constraint variations after variation, whereas the latter is obtained
by differentiating the independently truncated constraint values.
Define the constrained tangent matrix
\begin{equation*}
\begin{aligned}
\mathbf A(\mathbf c,\boldsymbol{\eta})
=
\frac{\partial}{\partial\mathbf c}
\left[
\mathbf R_0(\mathbf c)
+
\mathbf B(\mathbf c)\boldsymbol{\eta}
\right]
\in
\mathbb R^{(2n-1)\times(2n-1)}.
\end{aligned}
\end{equation*}
Using the analytic residual rates, the mechanical and displacement
residuals are stabilized simultaneously by imposing
\begin{equation}
\begin{aligned}
\mathbf A\dot{\mathbf c}
+
\mathbf B\dot{\boldsymbol{\eta}}
+
\gamma
\left(
\mathbf R_0
+
\mathbf B\boldsymbol{\eta}
\right)
&=
\mathbf0,
\\
\mathbf H\dot{\mathbf c}
-
\begin{bmatrix}
\dot{\overline{\Delta l}}_1
\\
\dot{\overline{\Delta l}}_2
\end{bmatrix}
+
\gamma\mathbf h
&=
\mathbf0,
\qquad
\gamma>0.
\end{aligned}
\label{eq:displacement-exponential-stabilization}
\end{equation}
Assuming that the bordered matrix below is nonsingular, rearranging
\eqref{eq:displacement-exponential-stabilization} yields
\begin{equation}
\begin{aligned}
\begin{bmatrix}
\dot{\mathbf c}
\\
\dot{\boldsymbol{\eta}}
\end{bmatrix}
=
\begin{bmatrix}
\mathbf A
&
\mathbf B
\\
\mathbf H
&
\mathbf0_{2\times2}
\end{bmatrix}^{-1}
\begin{bmatrix}
-\gamma
\left(
\mathbf R_0
+
\mathbf B\boldsymbol{\eta}
\right)
\\[1mm]
\begin{bmatrix}
\dot{\overline{\Delta l}}_1
\\
\dot{\overline{\Delta l}}_2
\end{bmatrix}
-
\gamma\mathbf h
\end{bmatrix}.
\end{aligned}
\label{eq:displacement-coefficient-rate-system}
\end{equation}
This nonsingularity condition defines a regular constrained branch and
necessarily requires
\(\operatorname{rank}\mathbf H(\mathbf c)
=
\operatorname{rank}\mathbf B(\mathbf c)
=
2\).
Both rates are propagated internally, while only \(\mathbf c\) is used
to reconstruct the robot centerline. The initial coefficients and auxiliary multipliers are obtained jointly
by enforcing
\begin{equation}
\begin{aligned}
\mathbf R_0(\mathbf c(0))
+
\mathbf B(\mathbf c(0))
\boldsymbol{\eta}(0)=
\mathbf0,\ \mathbf h(\mathbf c(0))
=
\mathbf0.
\end{aligned}
\label{eq:displacement-initial-alignment}
\end{equation}
Under this regularity condition, the
updated coefficients are substituted into
\eqref{eq:cartesian-polynomial} to recover the displacement-driven
equilibrium configuration.
Comparison of the tendon-dependent part of
\eqref{eq:total-potential} with
\eqref{eq:displacement-constrained-functional} shows that
\(-\eta_i\) is the corresponding tendon tension. Hence, physical
admissibility requires
\(\eta_i\leq0\), \(i=1,2\); a positive multiplier would require
compressive tendon action and therefore lies outside the present
formulation.

\section{Numerical Simulation and Model Validation}
\label{sec:numerical-validation}

The numerical study evaluates the Cartesian equilibrium-shape accuracy, distributed strain accuracy along the propagation path, residual control, and online computational performance of the proposed formulation against pointwise geometric
variable-strain method (GVS) equilibrium solves~\cite{mathew2025reduced}.

\subsection{Simulation Setup}
\label{subsec:simulation-setup}

\begin{table}[!t]
\centering
\caption{\scriptsize Simulation Parameters, Numerical Settings, and Validation Cases}
\label{tab:simulation-parameters}
\footnotesize
\setlength{\tabcolsep}{2pt}
\renewcommand{\arraystretch}{1.05}

\begin{tabular}{
@{}
p{0.33\columnwidth}
p{0.10\columnwidth}
p{0.42\columnwidth}
p{0.10\columnwidth}
@{}}
\toprule
\textbf{Parameter}
& \textbf{Symbol}
& \textbf{Value or profile}
& \textbf{Unit} \\
\midrule

Backbone length
& \(L\)
& \(0.40\)
& m \\

Young's modulus
& \(E\)
& \(2\times10^{9}\)
& Pa \\

Diameter profile
& \(D(s)\)
& Case-dependent
& m \\

Area profile
& \(A(s)\)
& \(\pi D^2(s)/4\)
& \(\mathrm{m^2}\) \\

Second-moment profile
& \(I(s)\)
& \(\pi D^4(s)/64\)
& \(\mathrm{m^4}\) \\

Tendon-routing diameter
& \(W(s)\)
& Case-dependent
& m \\
Mass density
& \(\rho\)
& 11970
& \(\mathrm{kg/m^3}\) \\
X-axis distributed load
& \(q_x(s)\)
& \(0\)
& \(\mathrm{N/m}\) \\

Y-axis distributed load
& \(q_y(s)\)
& \(-\rho gA(s)\)
& \(\mathrm{N/m}\) \\

Effective axial stiffness
& $K_A(s)$
& \( 0.007EA(s)\) for Cases 3, 4
& \(\mathrm{N}\) \\

\midrule

Polynomial order
& \(n\)
& 10
& -- \\

Taylor order
& \(P\)
& 6
& -- \\

Propagation step
& \(\Delta t\)
& \(5\times10^{-4}\)
& s \\

Path duration
& \(T\)
& 20
& s \\

Residual-correction rate
& \(\gamma\)
& 100
& \(\mathrm{s^{-1}}\) \\

Spatial integration nodes
& --
& Adaptive
& -- \\

Computing platform
& --
& Intel Core i9-14900HX
& -- \\




\bottomrule
\end{tabular}

\vspace{1.5mm}

\centering
\scriptsize
\setlength{\tabcolsep}{0.7pt}
\setlength{\medmuskip}{2mu}
\setlength{\thickmuskip}{2mu}
\renewcommand{\arraystretch}{1.15}
\begin{tabular*}{\columnwidth}{@{\extracolsep{\fill}}clll@{}}
\toprule
\textbf{Case}
& \makecell[tl]{\textbf{Primary}\\\textbf{feature}}
& $\boldsymbol{D(s)}\,[\mathrm{m}]$
& $\boldsymbol{W(s)}\,[\mathrm{m}]$ \\
\midrule
1
& \makecell[tl]{Varying\\routing}
& $D_0=0.004$
& $W_1(s)=0.020-0.015(s/L)^3$ \\
\addlinespace[2pt]
2
& \makecell[tl]{Nonuniform\\geometry}
& $D_1(s)=0.006-0.003(s/L)$
& $W_0=0.02$ \\
\addlinespace[2pt]
3
& \makecell[tl]{Axial\\deformation}
& $D_0=0.004$
& $W_0=0.02$\\
\addlinespace[2pt]
4
& \makecell[tl]{Combined\\effects}
& $D_1(s)=0.006-0.003(s/L)$
& $W_1(s)=0.020-0.015(s/L)^3$ \\
\bottomrule
\end{tabular*}
\vspace{1.5mm}

\centering
\scriptsize
\setlength{\tabcolsep}{0.7pt}
\setlength{\medmuskip}{2mu}
\setlength{\thickmuskip}{2mu}
\renewcommand{\arraystretch}{1.15}
\begin{tabular*}{\columnwidth}{@{\extracolsep{\fill}}clll@{}}
\toprule
\textbf{Case}
& \makecell[tl]{\textbf{Primary}\\\textbf{feature}}
& $\boldsymbol{F_1(t)}\,[N]$
& $\boldsymbol{F_2(t)}\,[N]$ \\
\midrule
1
& \makecell[tl]{Varying routing}
& $1.6t$
& $0$ \\
\addlinespace[2pt]
2
& \makecell[tl]{Nonuniform geometry}
& $2.6t$
& $1t$ \\
\addlinespace[2pt]
3
& \makecell[tl]{Axial deformation}
& $2.0t$
& $0.4t$\\
\addlinespace[2pt]
4
& \makecell[tl]{Combined effects}
& $2.0t$
& $0.4t$ \\
\bottomrule
\end{tabular*}
\end{table}
Table~\ref{tab:simulation-parameters} lists the four cases,
numerical settings, and tendon-force profiles over \(0\leq t\leq T\).
Both models use identical physical parameters, gravitational loading
\(q_y(s)=-\rho gA(s)\), force samples, and initial equilibria.
In Cases~3 and 4, \(K_A(s)=0.007EA(s)\) replaces \(EA(s)\) as the
effective axial stiffness. For the GVS reference, the strain field is approximated using fifth-order
polynomial basis functions, with a residual acceptance threshold of
\(10^{-7}\); each load step is warm-started from the converged solution of
the preceding step to improve numerical robustness.

\subsection{Numerical Results and Analysis}
\label{subsec:numerical-results}

\begin{figure*}[t]
\centering
\captionsetup[subfloat]{font=scriptsize}
\subfloat[\scriptsize Varying routing
\label{fig:shape-case-1}]
{\includegraphics[width=0.25\textwidth]
{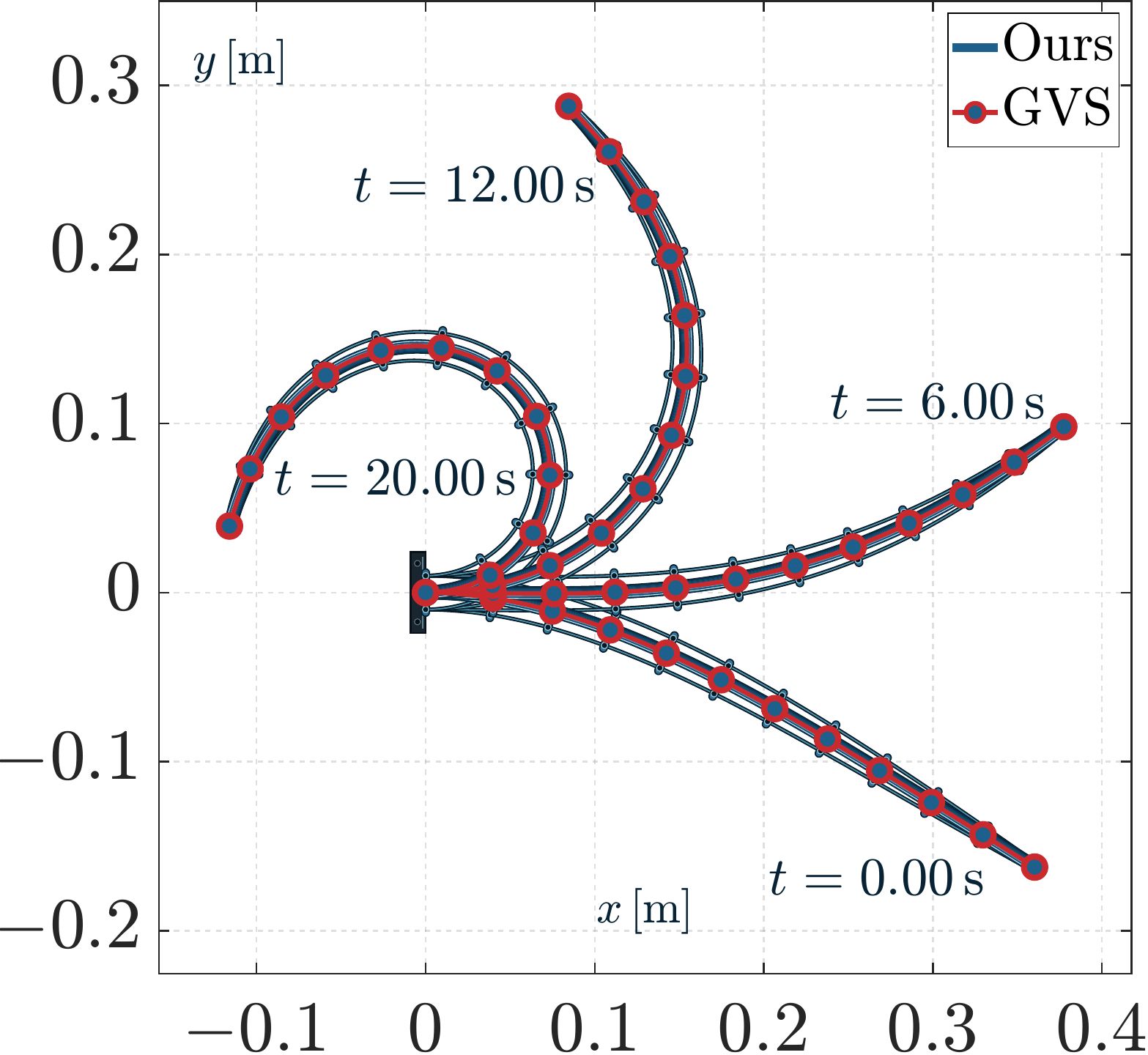}}
\hfill
\subfloat[\scriptsize Nonuniform geometry
\label{fig:shape-case-2}]
{\includegraphics[width=0.246\textwidth]
{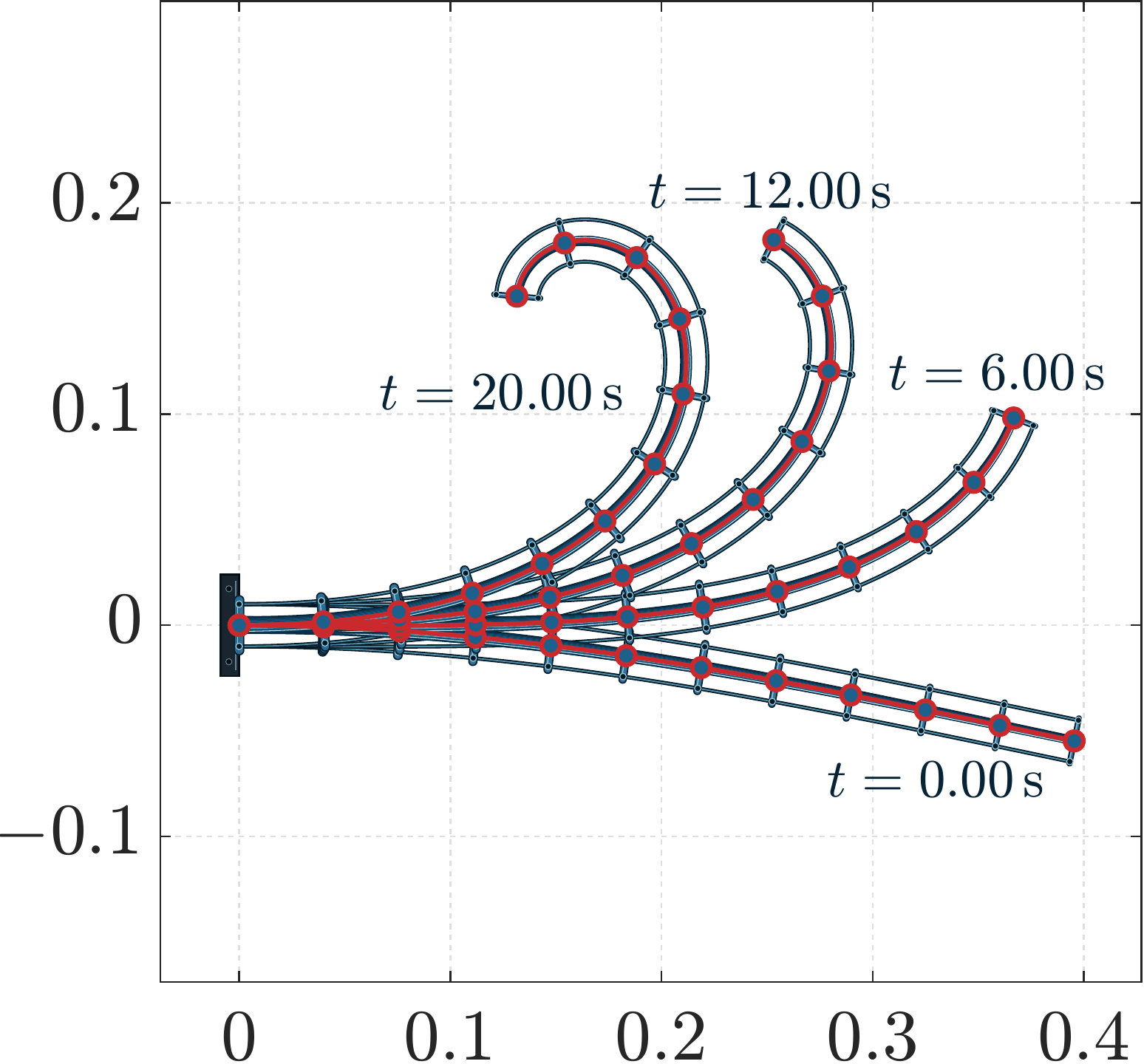}}
\hfill
\subfloat[\scriptsize Axial deformation
\label{fig:shape-case-3}]
{\includegraphics[width=0.246\textwidth]
{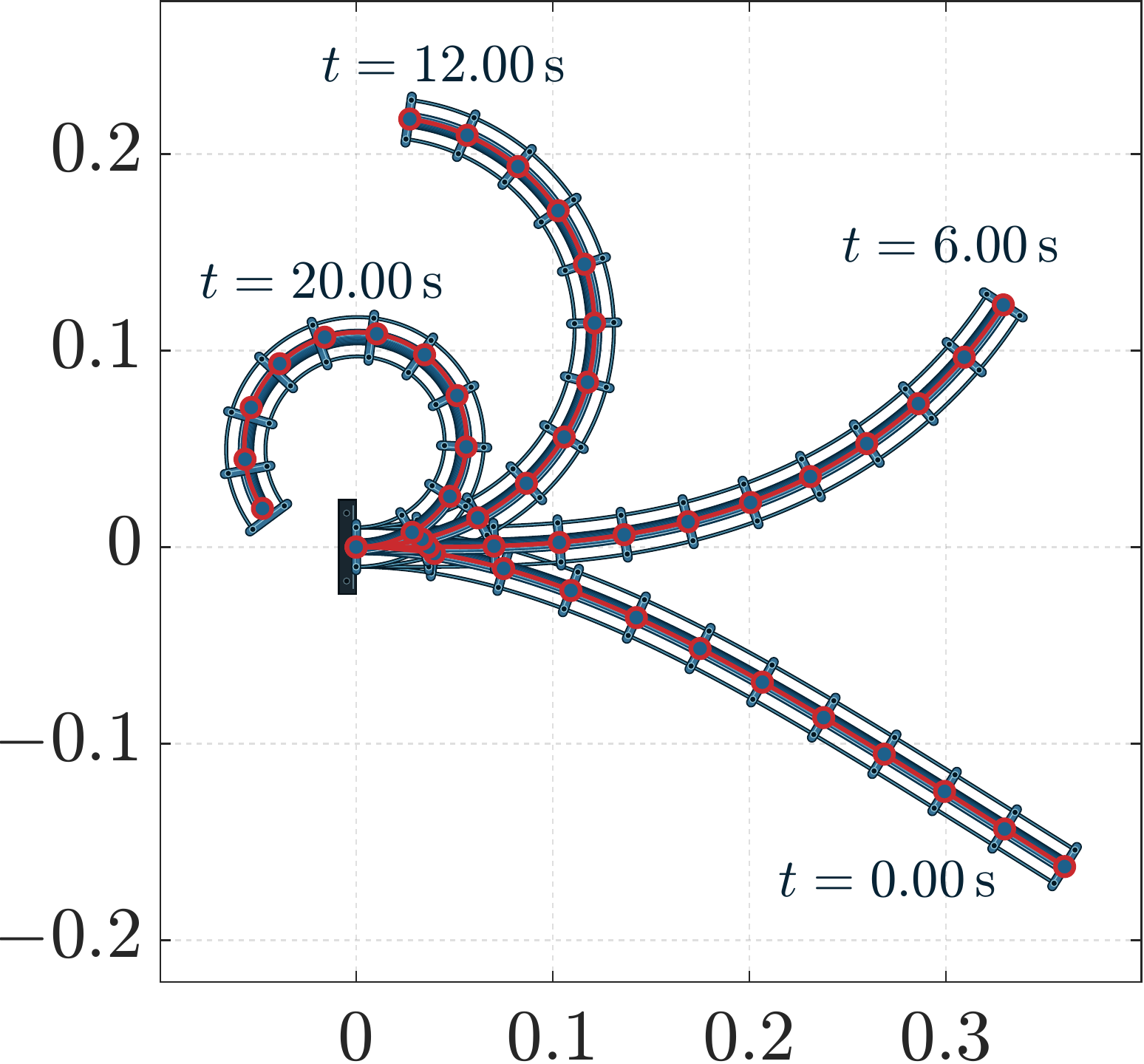}}
\hfill
\subfloat[\scriptsize Combined effects
\label{fig:shape-case-4}]
{\includegraphics[width=0.258\textwidth]
{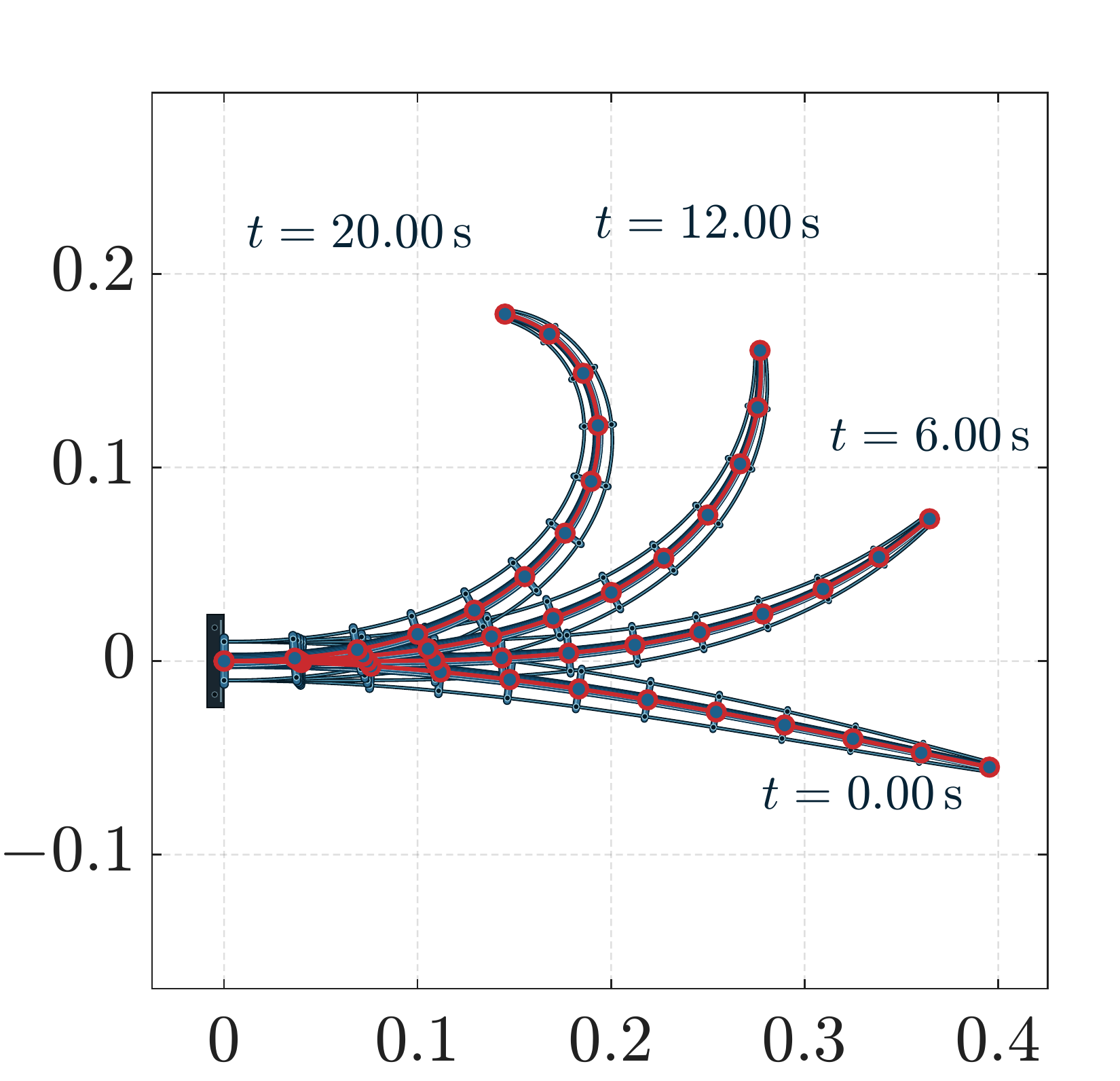}}

\par\vspace{-1.8mm}

\subfloat[Varying routing: \(e_\kappa\)
\label{fig:strain-error-case-1}]
{%
    \includegraphics[
        width=0.24\textwidth,
        clip
    ]{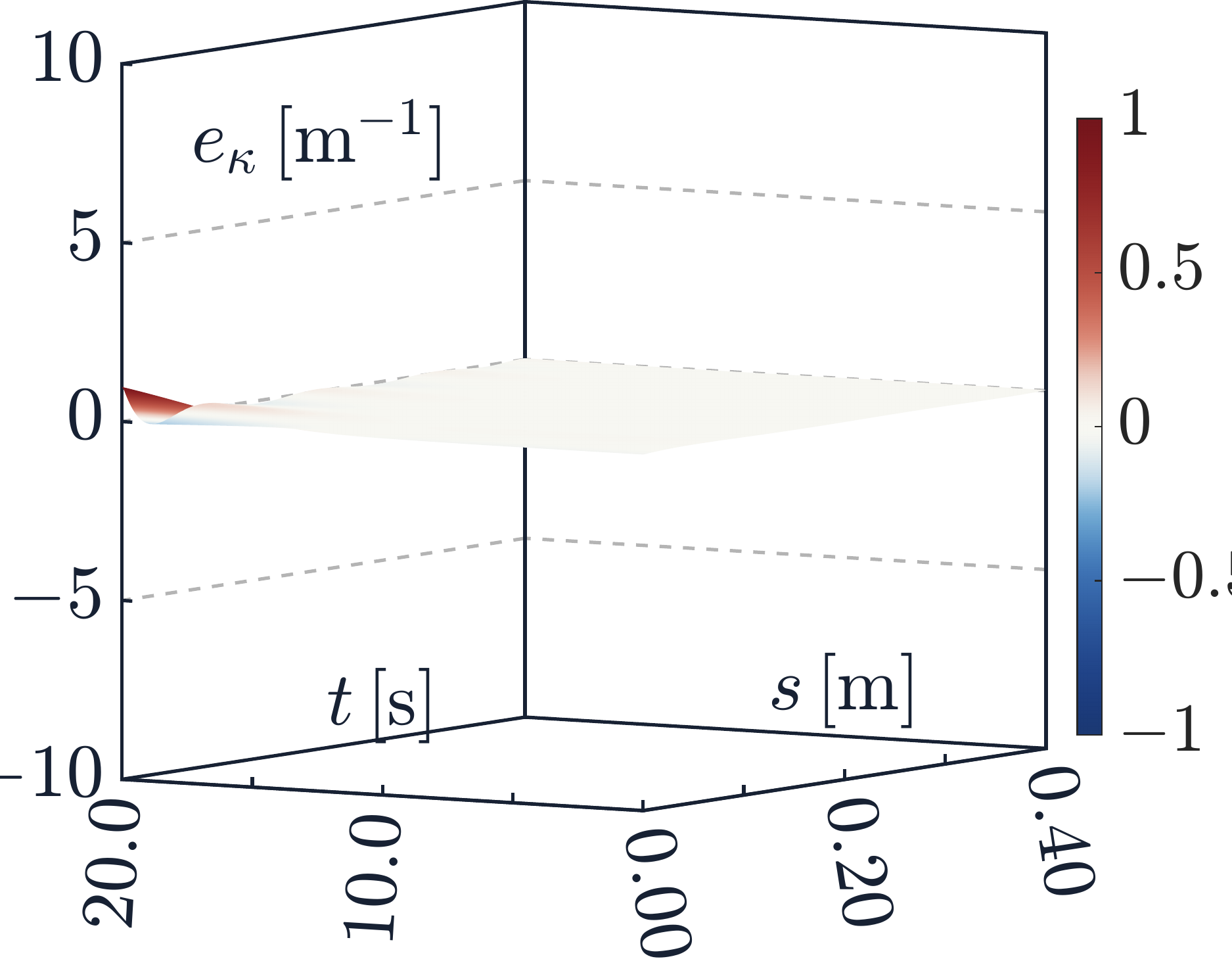}
}
\hfill
\subfloat[Nonuniform geometry: \(e_\kappa\)
\label{fig:strain-error-case-2}]
{%
    \includegraphics[
        width=0.24\textwidth,
        clip
    ]{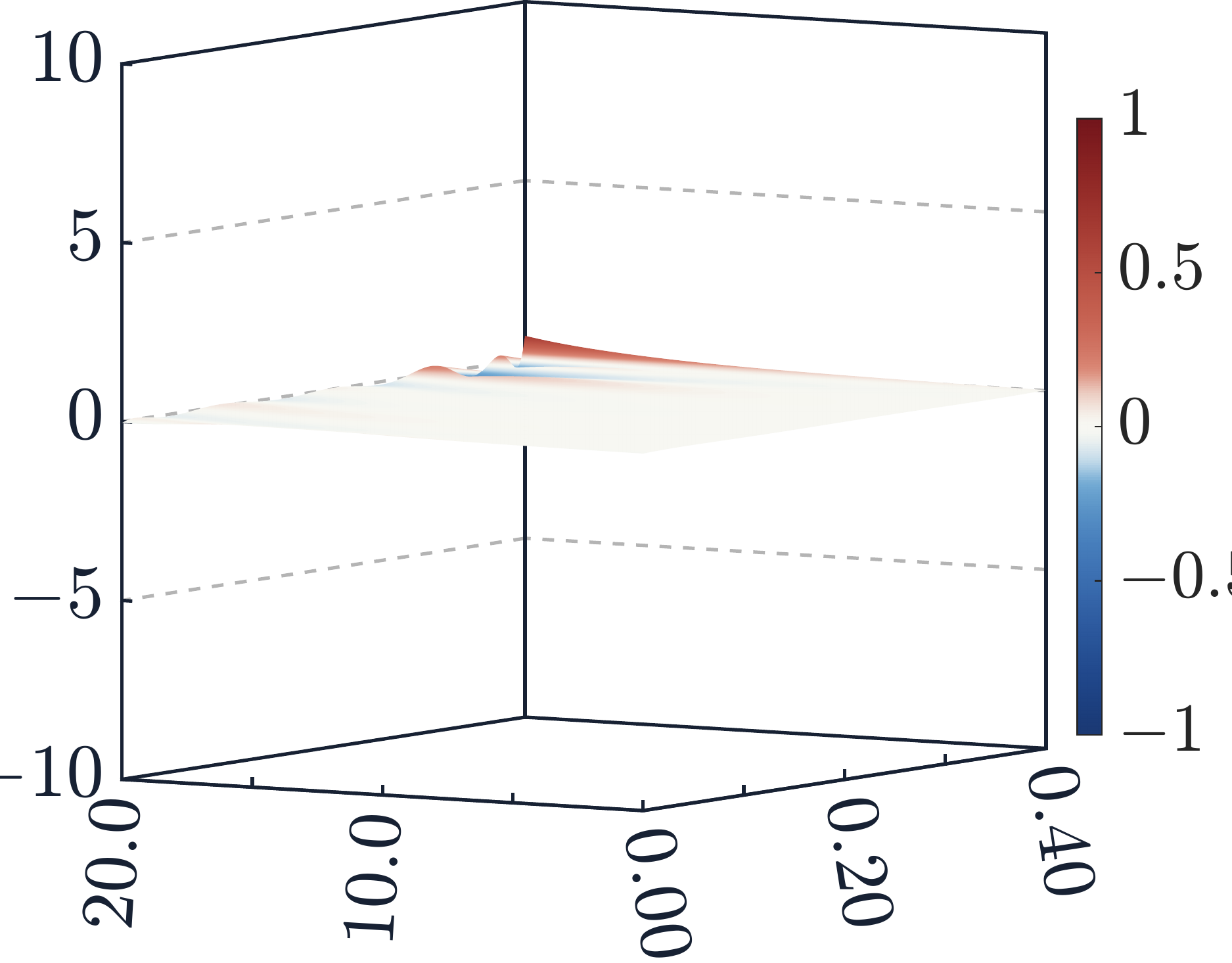}
}
\hfill
\subfloat[Axial deformation: \(e_\kappa,e_\lambda\)
\label{fig:strain-error-case-3}]
{%
    \begin{minipage}[b]{0.24\textwidth}
        \centering
        \includegraphics[width=\linewidth]{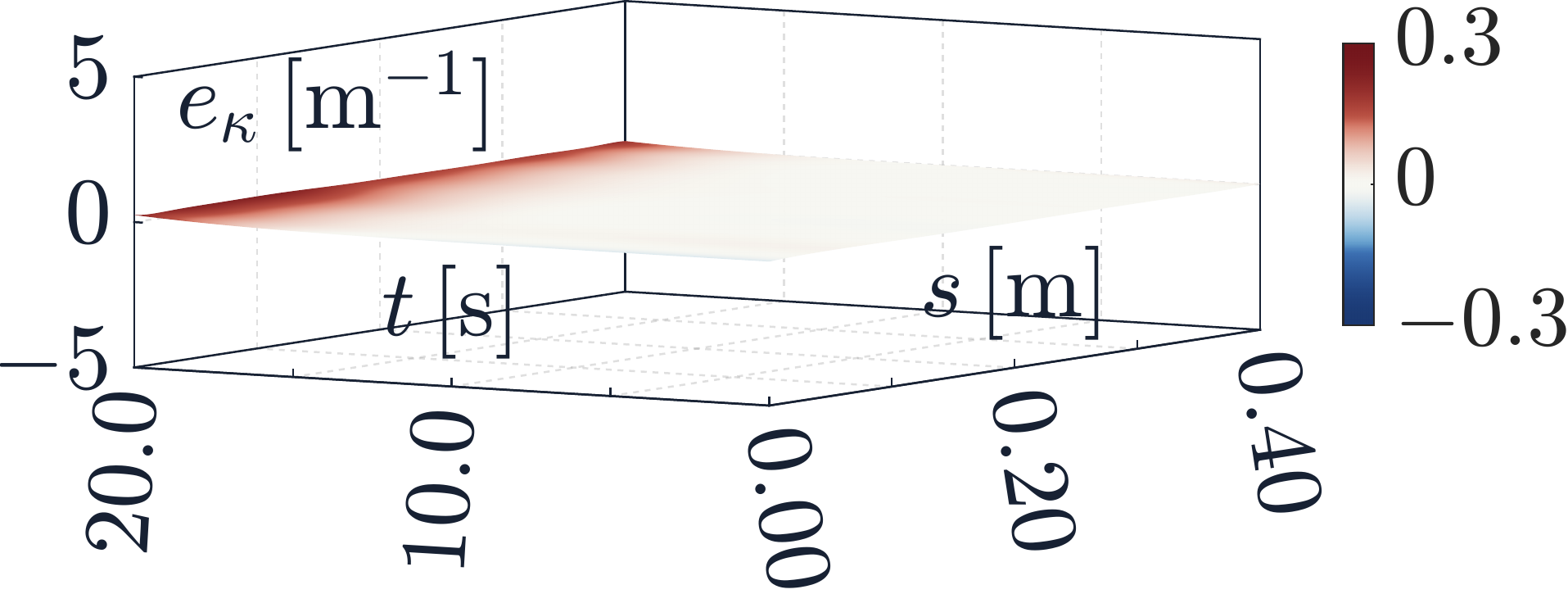}

        \vspace{-1mm}

        \includegraphics[width=\linewidth]{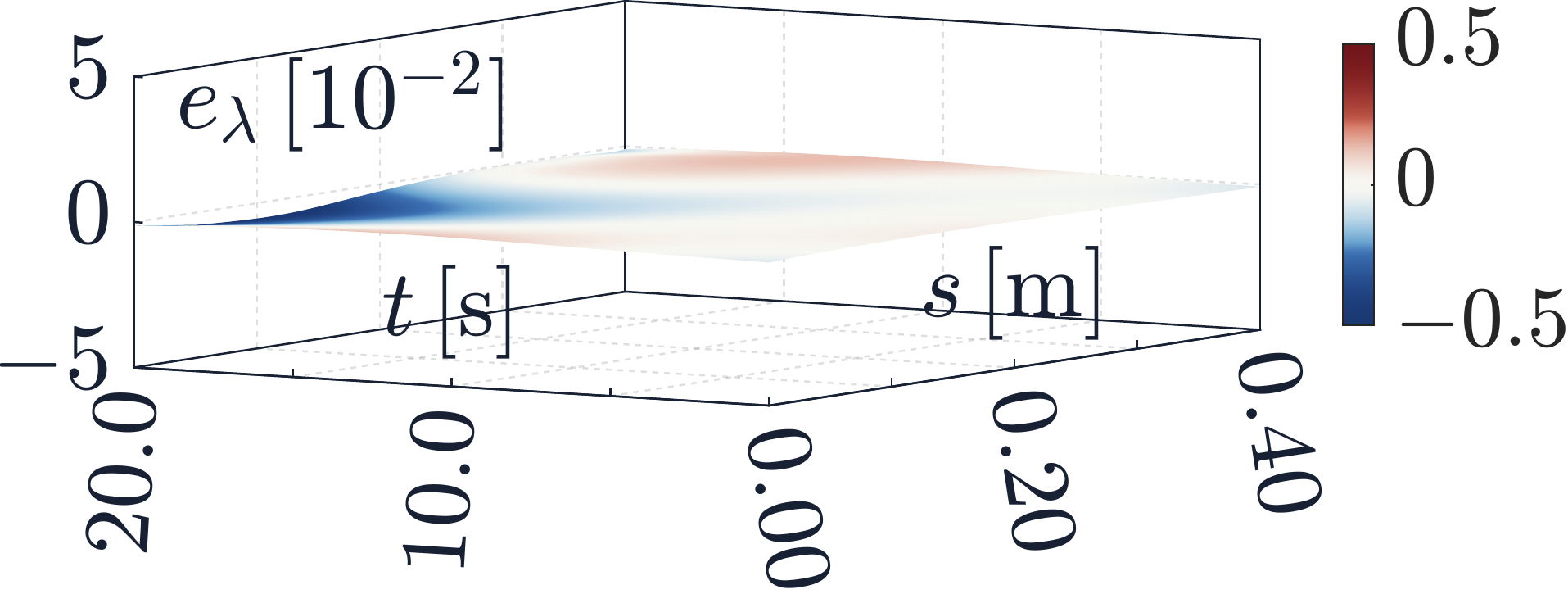}
    \end{minipage}
}
\hfill
\subfloat[Combined effects: \(e_\kappa,e_\lambda\)
\label{fig:strain-error-case-4}]
{%
    \begin{minipage}[b]{0.24\textwidth}
        \centering
        \includegraphics[width=\linewidth]{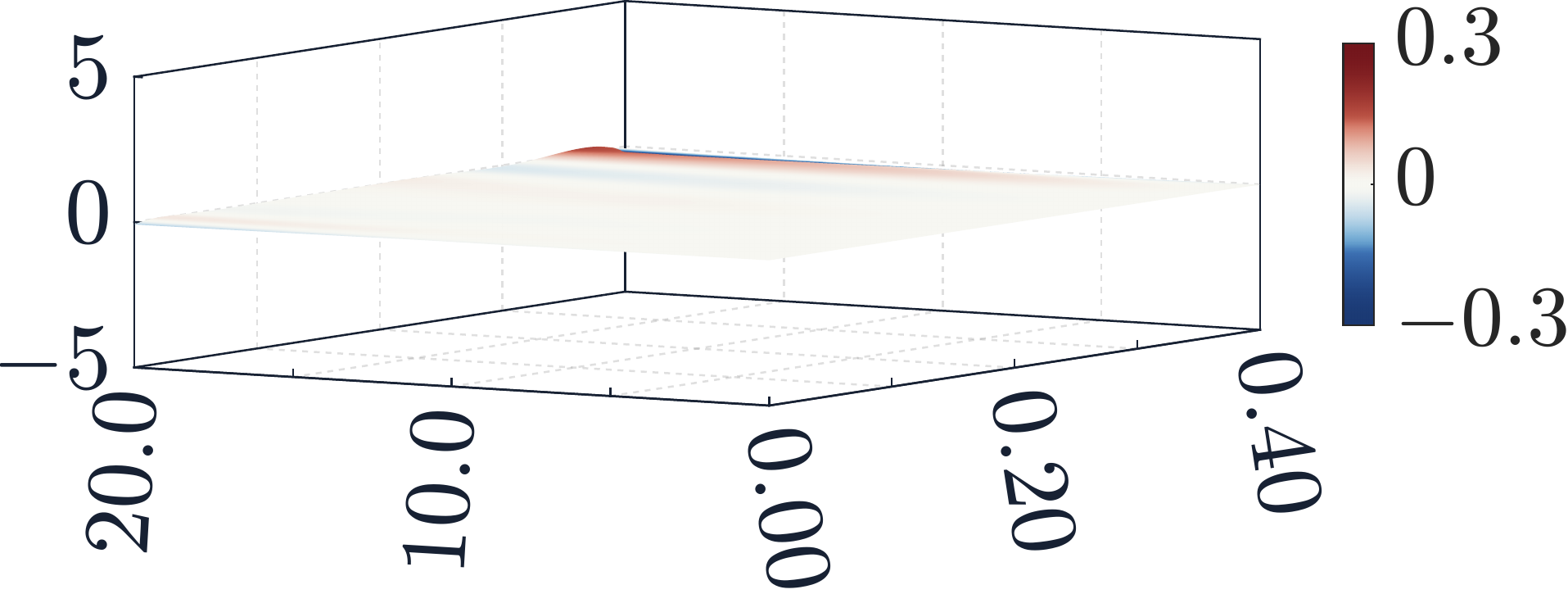}

        \vspace{-1mm}

        \includegraphics[width=\linewidth]{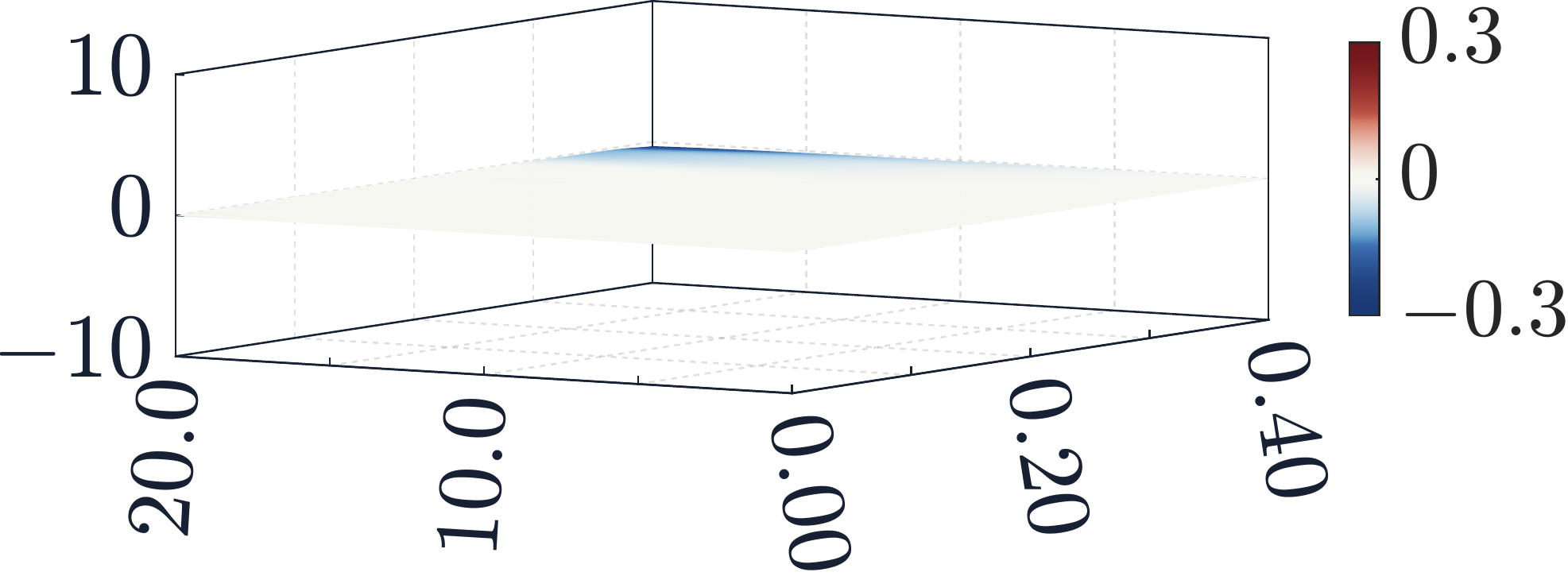}
    \end{minipage}
}


\par\vspace{-1.8mm}

\subfloat[Varying routing
\label{fig:residual-case-1}]
{\includegraphics[width=0.25\textwidth]
{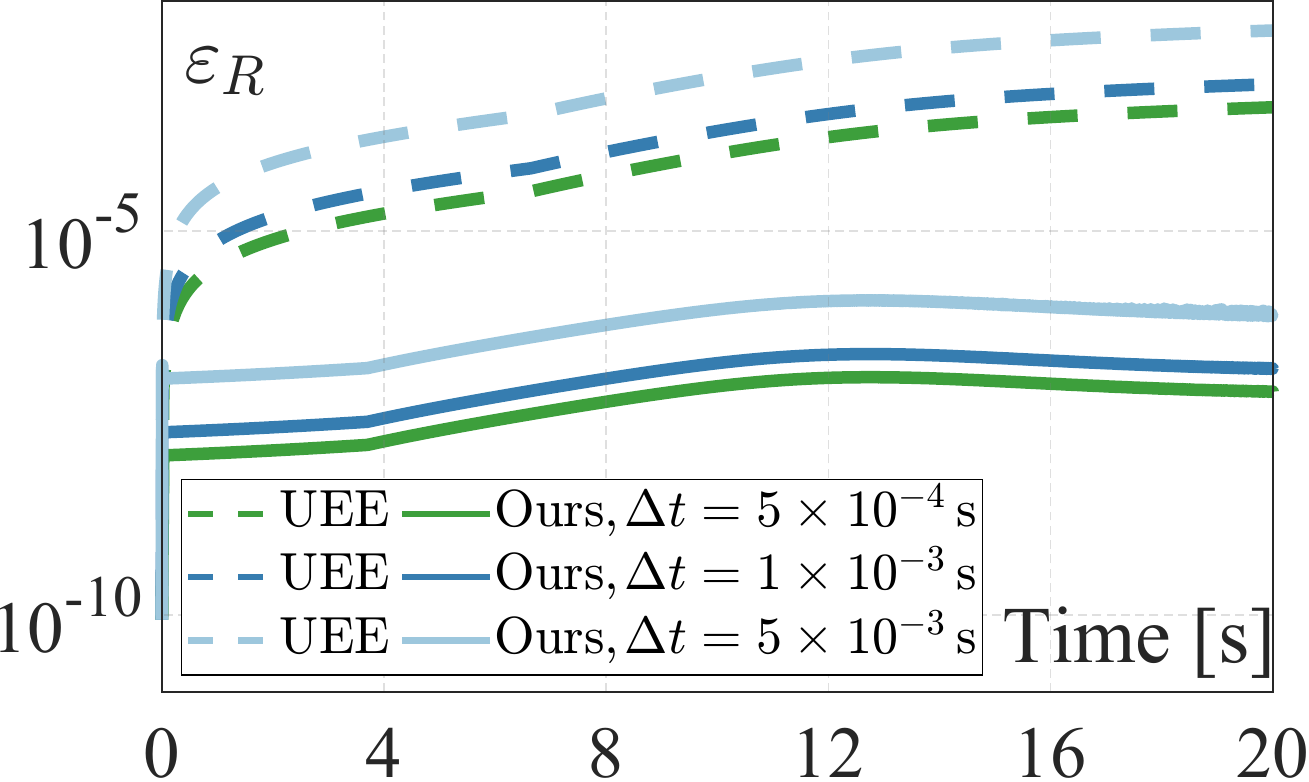}}
\hfill
\subfloat[Nonuniform geometry
\label{fig:residual-case-2}]
{\includegraphics[width=0.25\textwidth]
{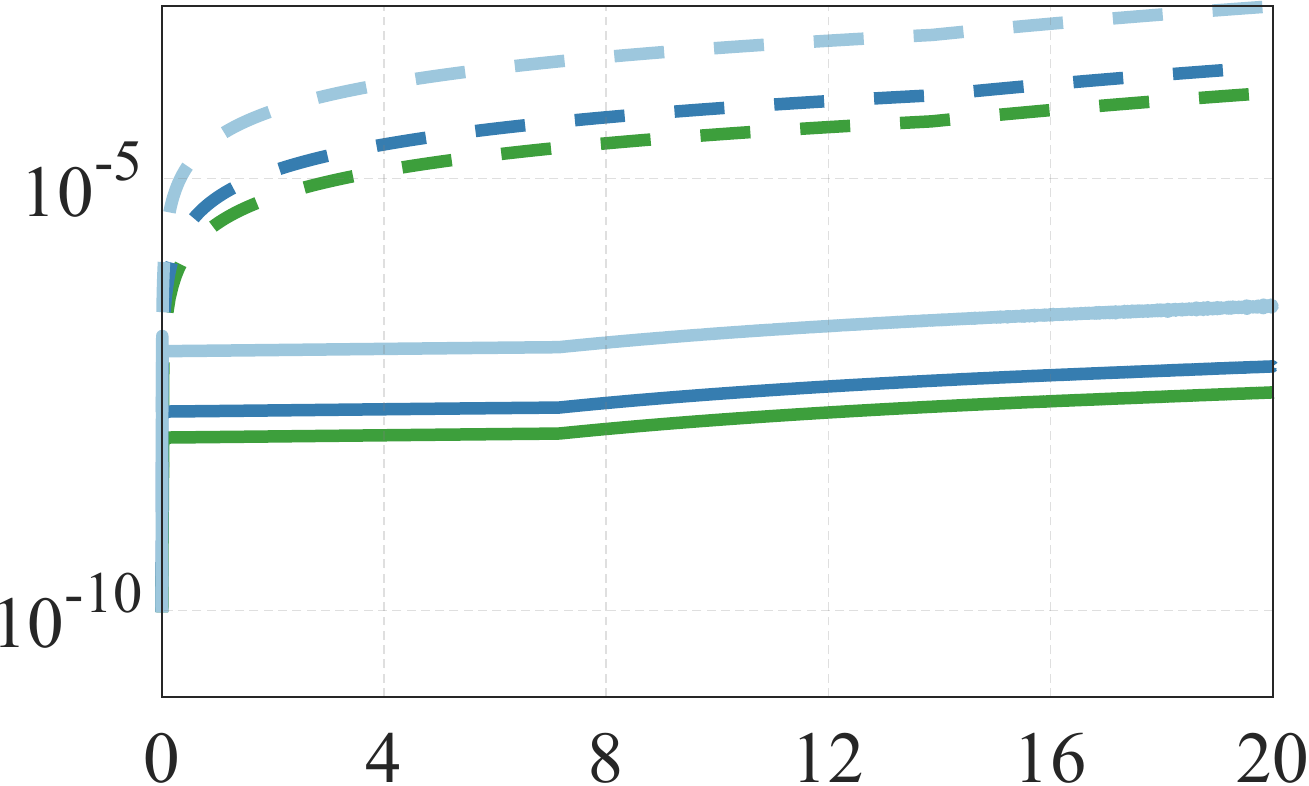}}
\hfill
\subfloat[Axial deformation
\label{fig:residual-case-3}]
{\includegraphics[width=0.25\textwidth]
{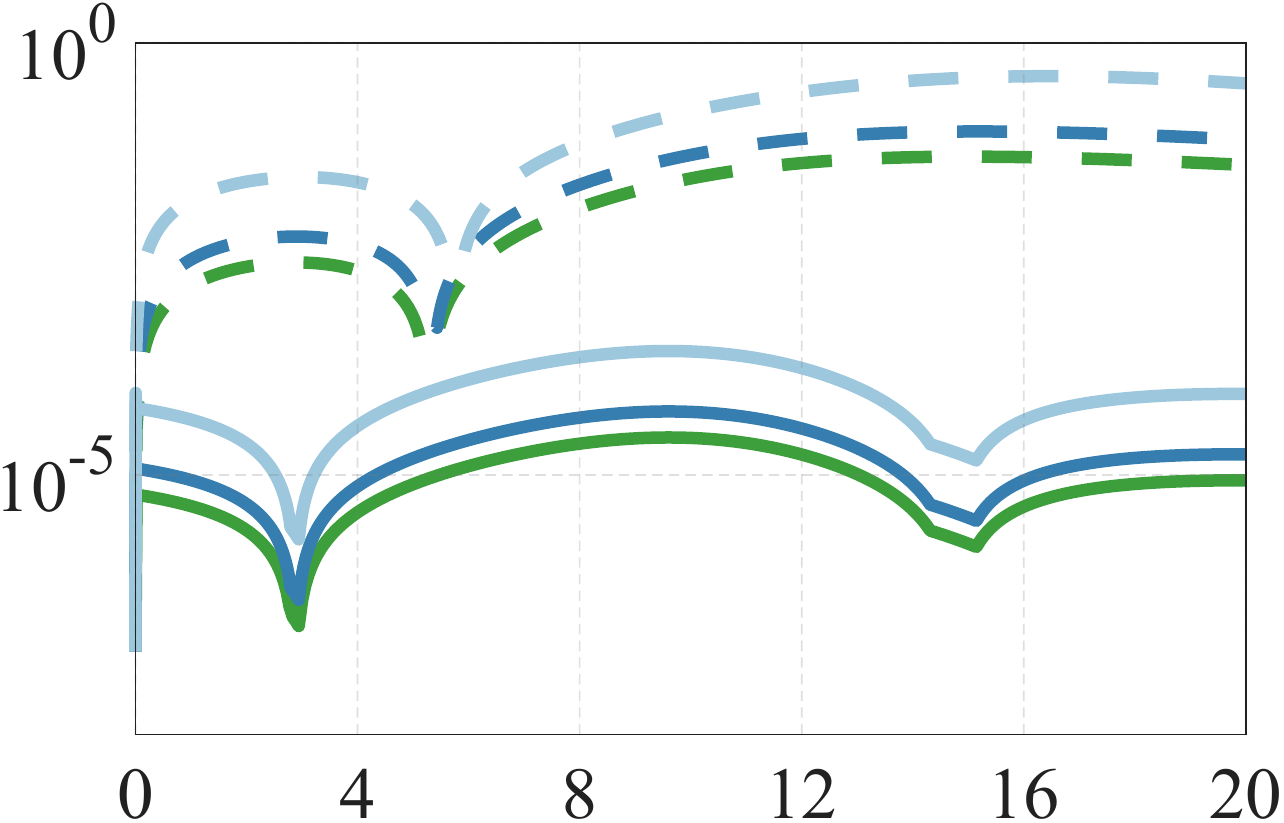}}
\hfill
\subfloat[Combined effects
\label{fig:residual-case-4}]
{\includegraphics[width=0.25\textwidth]
{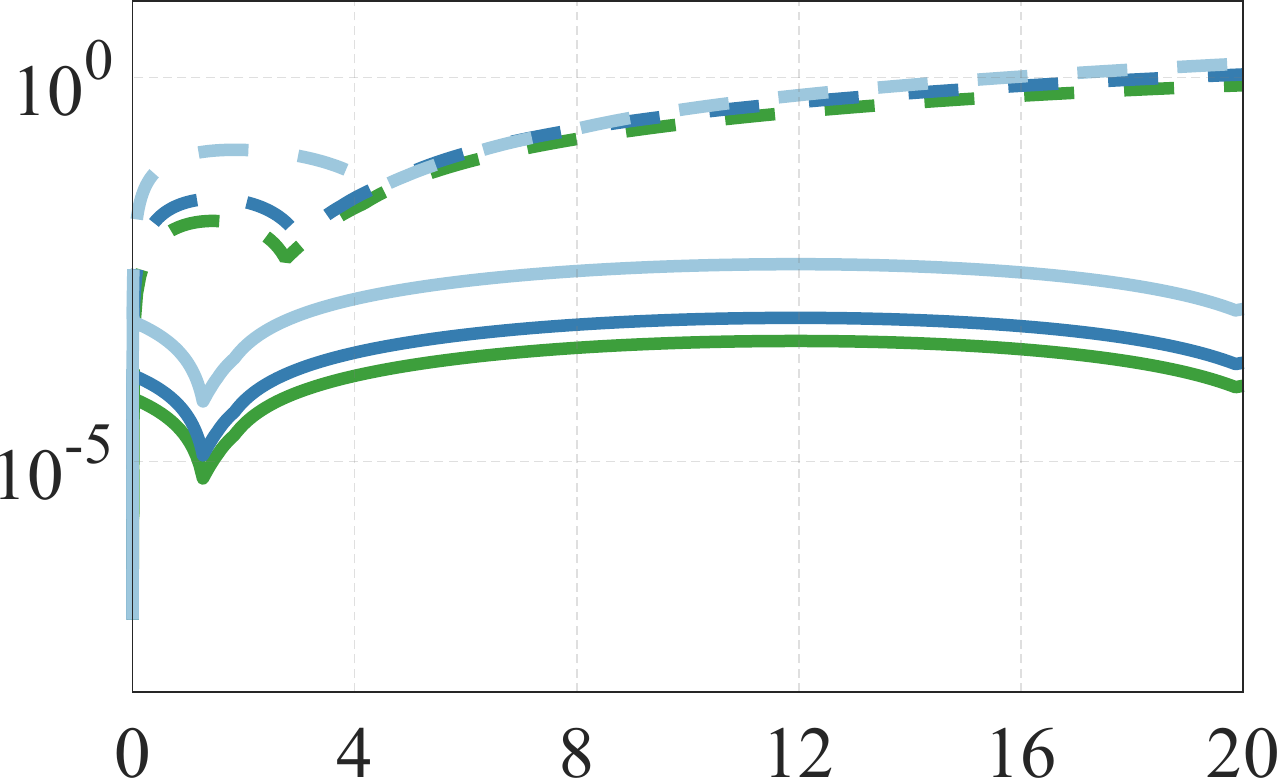}}

\caption{Numerical validation across the four mechanical cases.}
\label{fig:numerical-results}
\end{figure*}

\subsubsection{Cartesian Shape and Distributed Strain Accuracy}
\label{subsubsec:shape-accuracy}
The first row of Fig.~\ref{fig:numerical-results} shows close
agreement between the propagated Cartesian backbones and pointwise
GVS solutions across all four cases. The signed strain errors are
\(e_{\kappa}(s,t)=
\kappa_{\mathrm{prop}}(s,t)-\kappa_{\mathrm{GVS}}(s,t)\)
and
\(e_{\lambda}(s,t)=
\lambda_{\mathrm{prop}}(s,t)-\lambda_{\mathrm{GVS}}(s,t)\).
Their distributions along the backbone and force path appear in the
second row. The maximum absolute bending-strain errors
\(\max_{s,t}|e_\kappa(s,t)|\) in Cases~1--4 are
$0.969$, $0.662$, $0.244$, and
$0.221\,\mathrm{m}^{-1}$, respectively.
The maximum absolute axial-stretch-ratio errors
\(\max_{s,t}|e_\lambda(s,t)|\) in Cases~3 and 4 are
$5.303\times10^{-3}$ and $3.033\times10^{-3}$.
These results demonstrate that the proposed model also accurately captures the GVS baseline responses in strain space.

\subsubsection{Residual Control and Computational Performance}
\label{subsubsec:residual-computation}

Residual control and step-size sensitivity are evaluated using
$\Delta t\in
\{5.0\times10^{-4},1.0\times10^{-3},5.0\times10^{-3}\}\mathrm{s}$. At each propagation step, the proposed
residual-corrected propagation is compared with its uncorrected
explicit-Euler counterpart. Both
methods start from the same aligned initial equilibrium. The
equilibrium residual along each propagated trajectory is measured
directly as
$
\mathcal E_R(t;\Delta t)
=
\left\|
\mathbf R
\right\|_{\infty}.$
The third row of Fig.~\ref{fig:numerical-results} shows the resulting
equilibrium-residual histories for the tested propagation steps. Across the four cases, the proposed method suppresses accumulated equilibrium-residual drift and maintains substantially lower residuals across the tested step sizes, whereas uncorrected explicit Euler (UEE) exhibits larger, step-size-dependent residual drift during propagation. Let
\(\overline t_{\mathrm{upd},m}\) denote the average update time of method
\(m\). Its runtime relative to the proposed method is
$
\Gamma_m
=
\frac{\overline t_{\mathrm{upd},m}}
     {\overline t_{\mathrm{upd,prop}}},
$
where \(\Gamma_m>1\) indicates a longer update time than the proposed
method.
\begin{table}[!t]
\centering
\caption{\scriptsize Online Per-Update Computation Times and Relative Runtime Ratios}
\label{tab:computation-time}
\scriptsize
\setlength{\tabcolsep}{1.2pt}
\renewcommand{\arraystretch}{1.05}
\begin{tabular*}{\columnwidth}{@{\extracolsep{\fill}}lcccccc@{}}
\toprule
\textbf{Method}
& \textbf{Case 1}
& \textbf{Case 2}
& \textbf{Case 3}
& \textbf{Case 4}
& \textbf{Mean}
& $\boldsymbol{\Gamma_m}$ \\
\midrule
Proposed
& 0.107 & 0.153 & 0.332 & 1.440 & 0.508 & $1.000$ \\
GVS
& 5.347 & 4.160 & 4.790 & 8.045 & 5.586 & $10.996$ \\
Uncorrected Euler
& 0.107 & 0.153 & 0.328 & 1.425 & 0.503 & $0.990$ \\
\bottomrule
\end{tabular*}
\par\vspace{2pt}
\parbox{\columnwidth}{\scriptsize
\textit{Note:} All computation times are in milliseconds (ms);
$\Gamma_m$ is dimensionless.
}
\end{table}
As shown in Table~\ref{tab:computation-time}, the proposed method achieves a mean update time of $0.508\,\mathrm{ms}$, compared with $5.586\,\mathrm{ms}$ for pointwise GVS equilibrium solves, corresponding to an approximately $11$-fold acceleration. Residual correction adds only about $0.98\,\%$ to the mean update cost of uncorrected Euler, providing substantially improved equilibrium tracking with minimal computational overhead.

\section{Experimental Validation}
\label{sec:experimental-validation}

The experiment validates displacement-driven prediction of the
backbone shape and end-effector motion along a linear
tendon-displacement path, considering the complete backbone-shape trajectory, its distributed Cartesian error, and the corresponding end-effector trajectory.

\subsection{Experimental Setup}
\label{subsec:experimental-setup}
\begin{figure}[t]
\centering
\includegraphics[width=1\columnwidth]{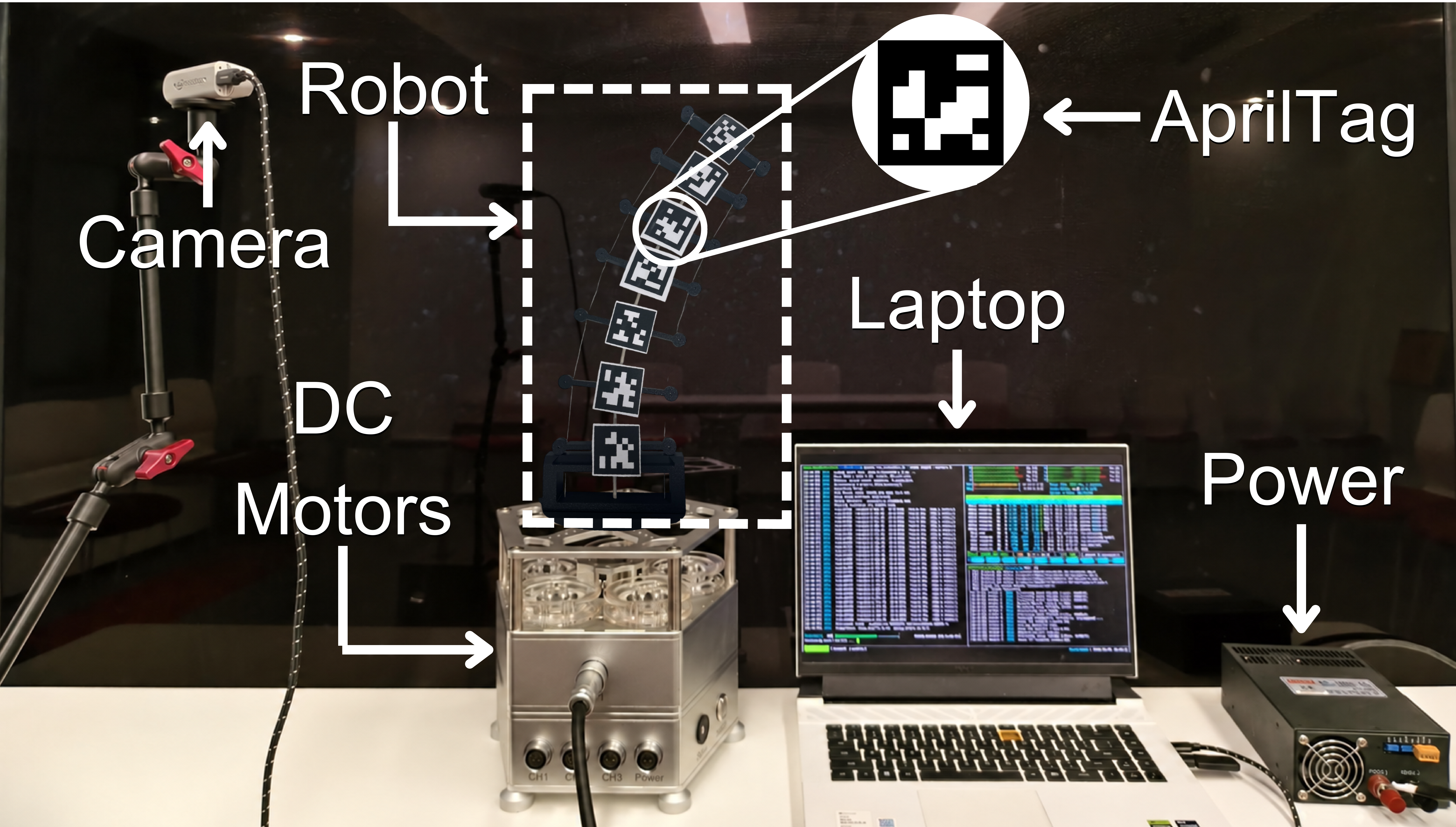}
\caption{Experimental platform.}
\label{fig:experimental-setup}
\end{figure}

The platform in Fig.~\ref{fig:experimental-setup} uses two
antagonistic motor-driven tendons and an RGB-D camera tracking seven
backbone markers. Table~\ref{tab:experimental-parameters} lists the
robot parameters and linear input profile. Axial deformation is
negligible over the tested range, with \(\lambda(s)\approx1\), and
the commanded tendon displacements satisfy
\(\overline{\Delta l}(t)=\overline{\Delta l_1}(t)
=-\overline{\Delta l_2}(t)\).

\begin{table}[!t]
\centering
\caption{Experimental Parameters and Tendon-Displacement Input}
\label{tab:experimental-parameters}
\footnotesize
\setlength{\tabcolsep}{2pt}
\renewcommand{\arraystretch}{1.08}
\begin{tabular}{
@{}
p{0.37\columnwidth}
p{0.14\columnwidth}
p{0.31\columnwidth}
p{0.10\columnwidth}
@{}}
\toprule
\textbf{Parameter}
& \textbf{Symbol}
& \textbf{Value or profile}
& \textbf{Unit} \\
\midrule

Backbone length
& \(L\)
& 0.4
& m \\

Cross-sectional diameter
& \(D\)
& 0.004
& m \\

Young's modulus
& \(E\)
& $2\times10^9$
& Pa \\

Tendon-routing diameter
& \(W\)
& 0.11
& m \\

Cross-sectional area
& \(A\)
& $1.26\times10^{-5}$
& \(\mathrm{m^2}\) \\

Second moment of area
& \(I\)
& $1.26\times10^{-11}$
& \(\mathrm{m^4}\) \\

Mass density
& \(\rho\)
& 11970
& \(\mathrm{kg/m^3}\) \\
\midrule

Tendon-displacement input
& \(\overline{\Delta l}(t)\)
& $\frac{80}{9} t, 0\le t\le 9$
& mm \\

\bottomrule
\end{tabular}
\end{table}

\subsection{Experimental Results and Analysis}
\label{subsec:experimental-results}

\begin{figure*}[t]
\centering
\captionsetup[subfloat]{font=scriptsize}
\subfloat[Backbone-shape trajectory
\label{fig:exp-backbone-trajectory}]
{\includegraphics[width=0.30\textwidth]
{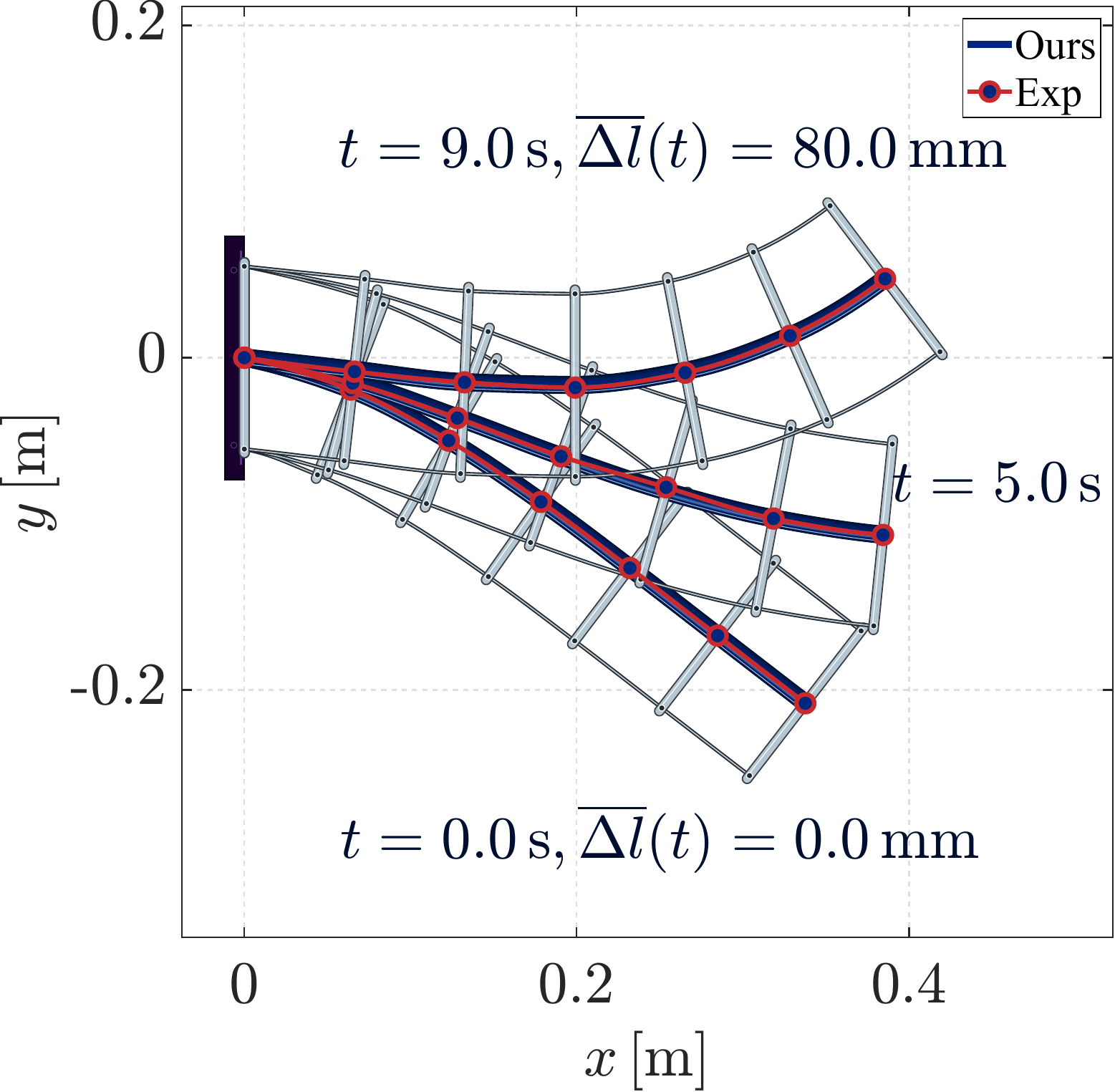}}
\hfill
\subfloat[Distributed Cartesian error
\label{fig:exp-backbone-error}]
{\includegraphics[width=0.38\textwidth]
{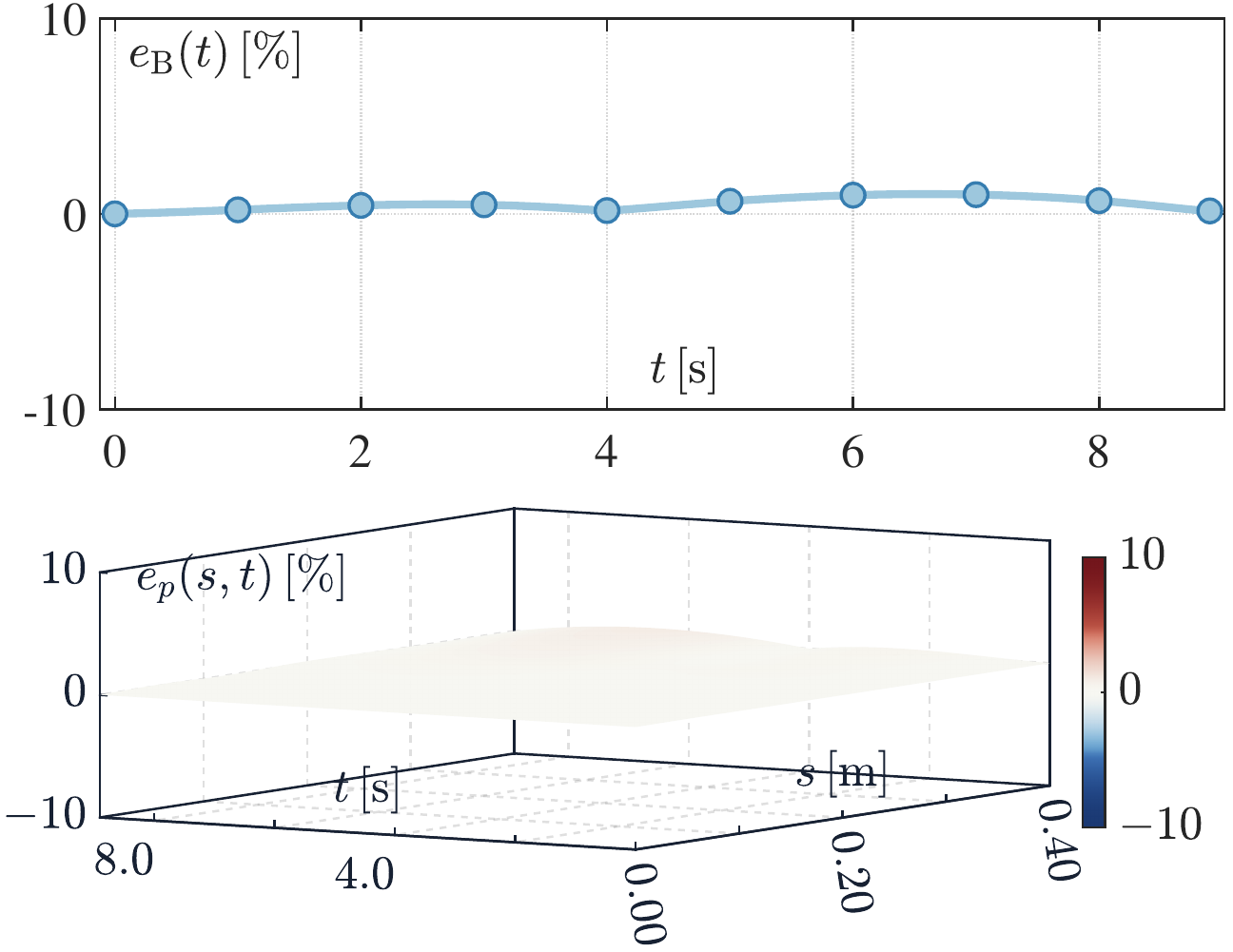}}
\hfill
\subfloat[End-effector trajectory
\label{fig:exp-tip}]
{\includegraphics[width=0.29\textwidth]
{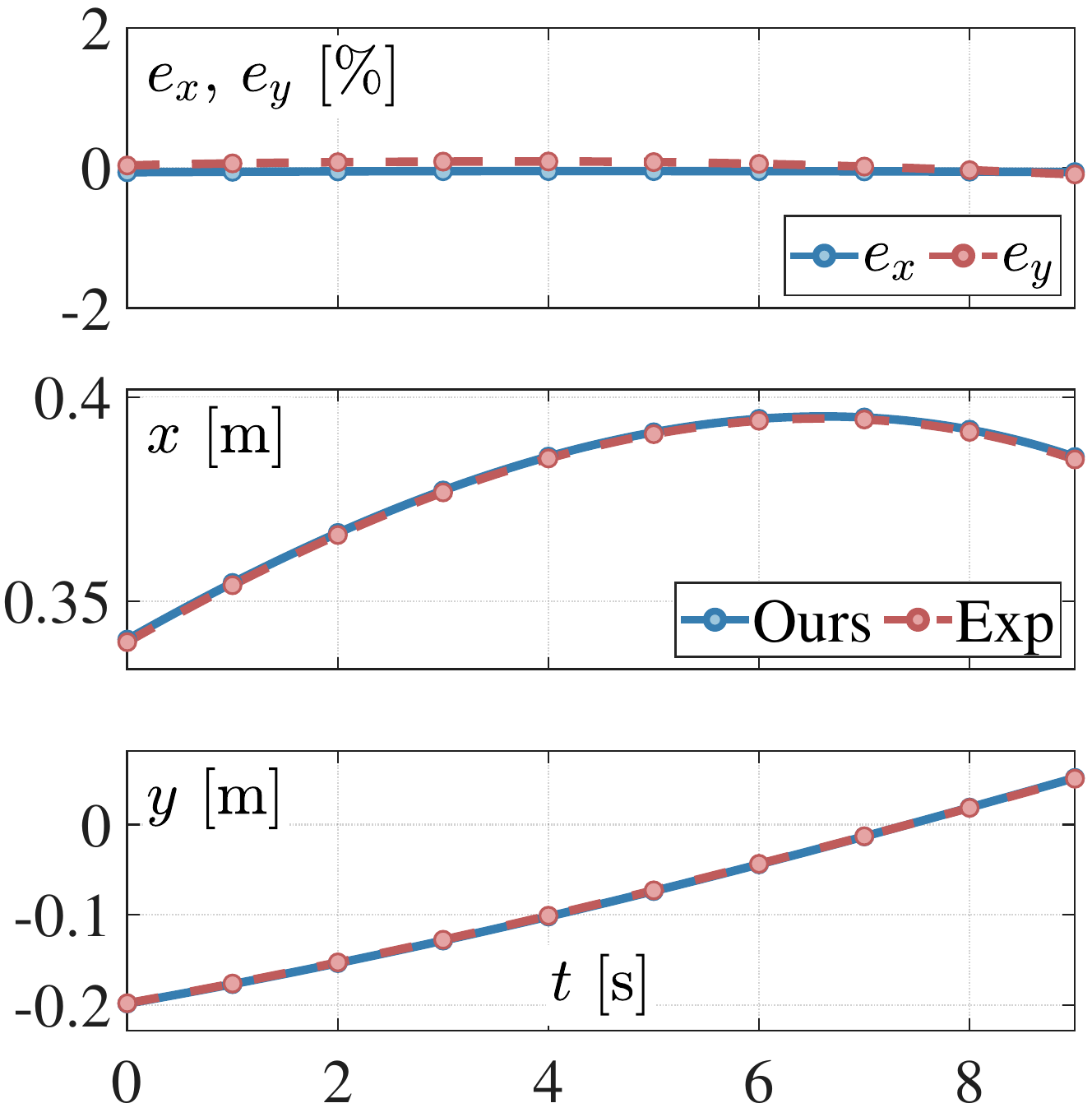}}

\caption{Experimental validation under the linear tendon-displacement
input.}
\label{fig:experimental-results}
\end{figure*}

Let $\mathbf{p}_{\mathrm{pred}}(s,t)$ and
$\mathbf{p}_{\mathrm{exp}}(s,t)$ denote the predicted
and measured backbone positions, respectively:
\begin{equation*}
\begin{aligned}
\mathbf{p}_{\mathrm{pred}}(s,t)
&= \bigl[x_{\mathrm{pred}}(s,t),\,
          y_{\mathrm{pred}}(s,t)\bigr]^{\mathrm{T}},\\
\mathbf{p}_{\mathrm{exp}}(s,t)
&= \bigl[x_{\mathrm{exp}}(s,t),\,
          y_{\mathrm{exp}}(s,t)\bigr]^{\mathrm{T}}.
\end{aligned}
\end{equation*}
Define the normalized pointwise error as
\(e_p(s,t)=
\|\mathbf p_{\mathrm{pred}}(s,t)
-\mathbf p_{\mathrm{exp}}(s,t)\|_2/L \times 100\%\),
the normalized mean backbone error as
\(e_{\mathrm B}(t)=
\int_0^L e_p(s,t)\,\mathrm ds/L \times 100\% \),
and the normalized Euclidean tip error as
\(e_x(t)=x_{\mathrm{pred}}(L,t)-x_{\mathrm{exp}}(L,t)\), \(e_y(t)=y_{\mathrm{pred}}(L,t)-y_{\mathrm{exp}}(L,t)\).
All position-error statistics are normalized by $L$ and reported
as percentages. Fig.~\ref{fig:experimental-results} compares the
backbone shapes, distributed errors, and end-effector trajectories.
The measured and predicted backbone shapes in
Fig.~\ref{fig:exp-backbone-trajectory} agree closely throughout the
actuation path. The maximum normalized mean backbone error is
\(\max_t e_{\mathrm B}(t)=1.02\%\), and the pointwise errors in
Fig.~\ref{fig:exp-backbone-error} remain small along the backbone.
The measured and predicted $x$- and $y$-tip trajectories in
Fig.~\ref{fig:exp-tip} also agree closely, with a maximum normalized
Euclidean error of \(\max_t |e_x(t)|=0.062\%, \max_t |e_y(t)|=0.092\%\).
These results support accurate Cartesian full-shape prediction
under tendon-displacement actuation.

\section{Conclusion}
\label{sec:conclusion}

This paper presents a reduced Cartesian formulation that propagates the complete equilibrium shape of planar TDCRs along prescribed actuation paths. Offline spatial moments and analytic residual derivatives enable force- and displacement-driven updates with one fixed-dimensional linear solve per rate evaluation after initial equilibrium alignment on a regular branch. Across four simulated mechanical cases, the propagated Cartesian configurations and distributed strains closely match GVS, while residual correction suppresses the drift observed with uncorrected Euler. The mean update time is $0.508\,\mathrm{ms}$, approximately $11$ times faster than pointwise GVS equilibrium solves, with only about $0.98\%$ additional cost relative to uncorrected Euler. Displacement-driven experiments yield a maximum normalized mean backbone position error of $1.02\,\%$ and maximum normalized end-effector errors of $0.062\%$ and $0.092\%$ in the x and y axes. Together, these results support efficient and accurate Cartesian full-shape prediction with controlled residual drift along the tested actuation paths. Future work will address three-dimensional deformation, tendon friction, and material hysteresis.

\bibliographystyle{IEEEtran}
\bibliography{reference}

\end{document}